\documentclass[journal]{IEEEtran}
\usepackage{amsmath,amsfonts}
\usepackage{algorithmic}
\usepackage{algorithm}
\usepackage{array}
\usepackage[caption=false,font=scriptsize,labelfont=sf,textfont=sf]{subfig}
\usepackage{textcomp}
\usepackage{stfloats}
\usepackage{url}
\usepackage{verbatim}
\usepackage{graphicx}
\usepackage{cite}
\usepackage{bm}
\usepackage{multirow}
\usepackage{threeparttable}
\usepackage{booktabs}
\usepackage{wrapfig}
\usepackage{tabularx}
\usepackage{etoolbox}
\makeatletter
\patchcmd{\@makecaption}
{\scshape}
{}
{}
{}
\makeatother

\begin{document}

\title{DNAD: Differentiable Neural \\ Architecture Distillation}

\author{Xuan Rao, Bo Zhao,~\IEEEmembership{Senior Member,~IEEE}, Derong Liu,~\IEEEmembership{Fellow,~IEEE}
\thanks{This work was supported in part by the National Natural Science Foundation of China under Grant 61973330, 62073085 and 62350055, in part by the Open Research Project of the Key Laboratory of Industrial Internet of Things \& Networked Control, Ministry of Education under Grant 2021FF10, in part by the Shenzhen Science and Technology Program under grant JCYJ20230807093513027, in part by the Fundamental Research Funds for the Central Universities under grant 1243300008, and in part by the Beijing Normal University Tang Scholar \emph{(Corresponding author: Bo Zhao)}.}
\thanks{Xuan Rao and Bo Zhao are with the School of Systems Science, Beijing Normal University, Beijing 100875, China (e-mail: raoxuan98@mail.bnu.edu.cn; zhaobo@bnu.edu.cn).}
\thanks{Derong Liu is with the School of Automation and Intelligent Manufacturing, Southern University of Science and Technology, Shenzhen 518000, China (e-mail: liudr@sustech.edu.cn), and also with the Department of Electrical and Computer Engineering, University of Illinois Chicago, Chicago, IL 60607, USA (e-mail: derong@uic.edu).}

	\thanks{© 2025 IEEE. This manuscript is submitted to IEEE Transaction on Neural Networks and Learning Systems. Personal use of this manuscript is permitted. Permission from IEEE must be obtained for all other uses, in any current or future media, including reprinting/republishing this material for advertising or promotional purposes, creating new collective works, for resale or redistribution to servers or lists, or reuse of any copyrighted component of this work in other works.}
	
}

\markboth{IEEE transaction on neural networks and learning systems}%
{Rao \MakeLowercase{\textit{et al.}}: Neural Architecture Distillation: Designing Resource-Aware Neural Network via Differentiable Feature Distillation and Super-Network Progressive Pruning}


\maketitle

\begin{abstract}
To meet the demand for designing efficient neural networks with appropriate trade-offs between model performance (e.g., classification accuracy) and computational complexity, the differentiable neural architecture distillation (DNAD) algorithm is developed based on two cores, namely \emph{search by deleting} and \emph{search by imitating}. Primarily, to derive neural architectures in a space where cells of the same type no longer share the same topology, the super-network progressive shrinking (SNPS) algorithm is developed based on the framework of differentiable architecture search (DARTS), i.e., \emph{search by deleting}. Unlike conventional DARTS-based approaches which yield neural architectures with simple structures and derive only one architecture during the search procedure, SNPS is able to derive a Pareto-optimal set of architectures with flexible structures by forcing the dynamic super-network shrink from a dense structure to a sparse one progressively. Furthermore, since knowledge distillation (KD) has shown great effectiveness to train a compact network with the assistance of an over-parameterized model, we integrate SNPS with KD to formulate the DNAD algorithm, i.e., \emph{search by imitating}. By minimizing behavioral differences between the super-network and teacher network, the over-fitting of one-level DARTS is avoided and well-performed neural architectures are derived. Experiments on CIFAR-10 and ImageNet classification tasks demonstrate that both SNPS and DNAD are able to derive a set of architectures which achieve similar or lower error rates with fewer parameters and FLOPs. Particularly, DNAD achieves the top-1 error rate of 23.7\% on ImageNet classification with a model of 6.0M parameters and 598M FLOPs, which outperforms most DARTS-based methods.    
\end{abstract}

\section{Introduction}
During the last decade, neural networks (NNs) and deep learning have witnessed tremendous progress in many artificial intelligence (AI) fields including computer vision, natural language processing, art-generation, etc. The representation ability of an NN is generally linked to its capacity, indicated by factors such as the number of parameters and floating point operations (FLOPs). Presently, NN models are growing in computational complexity to meet the rising demand for more powerful representation ability. However, the computational demands of large NNs pose challenges for deployment on resource-constrained devices, necessitating designing NN models with low power consumption, minimal memory footprint, and swift training. Note that these characteristics are crucial for real-time decision-making in applications such as auto-pilot and robot control.

As a prominent sub-topic within automated machine learning (Auto-ML), neural architecture search (NAS) \cite{he2021automl} has gained substantial attraction for its ability to automate the design process of NN architectures \cite{liu2018darts, xu2019pc, chen2019progressive, chen2020stabilizing, WANG2023109193, kang2020towards}. Many NAS approaches are developed based on reinforcement learning (RL) \cite{zoph2018learning, zhang2021adaptive, zhou2022automated}, evolutionary algorithm (EA) \cite{9723446, 10059145}, and Bayesian optimization \cite{kandasamy2018neural, 10753099}. However, these methods typically require extensive computational resources. 

Notably, differentiable architecture search (DARTS) has significantly improved the NAS efficiency by transforming the discrete search space into a continuous one \cite{liu2018darts, xu2019pc, chen2019progressive, chen2020stabilizing, WANG2023109193}. Recent advances in DARTS have sought to address its limitations in some degree. For instance, \cite{DBLP:conf/cvpr/YeL00FO22} proposed the $\beta$-decay regularization to stabilize the search process by constraining the magnitudes of architecture parameters. Similarly, \cite{9720178} introduced the cyclic optimization between search and evaluation networks via introspective distillation. To improve the robustness of DARTS, \cite{10672551} explored the adaptive channel allocation strategies to balance the computational efficiency and model performance. Meanwhile, \cite{10018939} developed a two-stage gradient-enhanced framework by combining the gradient-based block design with evolutionary network-level optimization. 

Recent advances have also explored integrating knowledge distillation (KD) with NAS to enhance architecture search. For instance, DBNAS employs deep supervision and attention with ground-truth labels for block-wise NAS \cite{yang2024deeply}, while DistillDARTS combines feature-based KD with DARTS to smooth the gradient distribution during the search process \cite{DBLP:conf/icmcs/LiaoZWYLRYF22}. In addition, KL-DNAS incorporates latent-aware objectives with distillation for efficient video processing \cite{singh2024kl}, and PAMKD addresses the collective failures by facilitating knowledge transfer among architectures \cite{xie2022performance}. Other methods leverage contrastive KD and NAS to guide the channel reduction \cite{10018843} and enhance the graph neural networks utilizing KD and NAS for scalable learning \cite{10328657}. However, these approaches always focus on specific sub-tasks (e.g., block-wise search or latency optimization), operate within topology-constrained search spaces, or rely on inefficient bi-level optimization, which limits their applicability.

Nevertheless, some drawbacks of DARTS-based approaches should be considered and solved. (1) Most methods are developed based on an over-simple search space where the whole neural architectures are constructed by repeatedly stacking cells of the same structure. It prevents us from obtaining more powerful neural networks by seeking more flexible architectures. (2) Conventionally, these methods can only derive one architecture in a single NAS procedure, which contradicts the demand for selecting neural architectures with different trade-offs between model performance and computational complexity. (3) When the one-level optimization is employed for DARTS, the super-network suffers from over-fitting and awful architectures can be derived \cite{liu2018darts}. GOLD-NAS \cite{bi2020gold} has attempted to solve the above drawbacks by pruning the over-sized super-network and claims that the collapse of one-level DARTS can be alleviated by some data-augment techniques. However, its implementation is time-consuming and the data-augmentation strategy is not powerful enough to improve the one-level DARTS, which is verified empirically in this paper. 

To further mitigate the above challenges, this paper develops the differentiable neural architecture distillation (DNAD) algorithm based on two cores, namely \emph{search by deleting} and \emph{search by imitating}.

\emph{Search by deleting} for problems (1) and (2). We inherit the inspirations from GOLD-NAS to explore architectures in an unconstrained space where networks are constructed by stacking neural cells of diverse structures. 
The reduction of prior rules is consistent with the goal of Auto-ML since the learning systems are with more freedom. However, deriving competitive architectures in this space is not an easy task, thus the super-network progressive shrinking (SNPS) algorithm is developed. Unlike DARTS approaches which utilize a sharp discretization strategy, SNPS compresses the super-network uninterruptedly from an initially dense structure to a progressively sparse one by pruning operators with small attention weights. In this way, SNPS derives architecture with diverse computational complexity in a single NAS procedure. In addition, the shrinking speed of the dynamic super-network is controlled by an adaptive performance-attentive sparsity entropy, which is, from the perspective of control theory, amended based on the difference between the expected pruning state and the actual state. 

\emph{Search by imitating} for problem (3). To improve the effectiveness of one-level optimization of DARTS, we regularize the optimization direction of the super-network by integrating SNPS with KD. 
The combination of SNPS and KD methods formulates the proposed DNAD algorithm, in which partial goal of NAS has turned into designing neural architectures which well imitate the teacher's behaviors. The main inspiration is that knowledge of a large network contains the reason why it generalizes well on unseen data, and the student dynamic super-network prevents itself from being over-fitting by behavior imitation. Specifically, intermediate feature maps of the teacher are utilized as additional supervised signals for the super-network. In DNAD, the feature maps of the teacher and super-network in different depths are collected and the $L_{2}$ losses corresponding to these feature maps are calculated then propagate along the parameters of the super-network. 
The contributions of this paper are summarized as follows.
\begin{itemize}{}{}
	\item{The DNAD algorithm is developed based on two cores, namely \emph{search by deleting} and \emph{search by imitating}.} 
	\item{In \emph{search by deleting}, the SNPS algorithm is developed under the framework of DARTS. In particular, SNPS searches neural architectures in a search space where cells of the same type no longer share the same topology. Furthermore, SNPS is able to derive a Pareto-optimal set of neural architectures with different computational complexity during one search procedure, which differs from most DARTS-based approaches that derive only one architecture in one search procedure.} 
	\item{In \emph{search by imitating}, SNPS is combined with KD, and the DNAD algorithm is derived correspondingly to improve the performance of searched neural architectures and stabilize the one-level optimization of DARTS.}
	\item{Experiments on CIFAR-10 and ImageNet demonstrate the effectiveness of the developed SNPS and DNAD. Both of them are able to achieve a superior tradeoff between the model performance and the computational complexity. In particular, the lowest error rates achieved by SNPS and DNAD on ImageNet are $23.9\%$ and $23.7\%$, which are lower than the state-of-the-art NAS baselines. }
\end{itemize}

\begin{figure}
	\centerline{\includegraphics[width=0.50\textwidth]{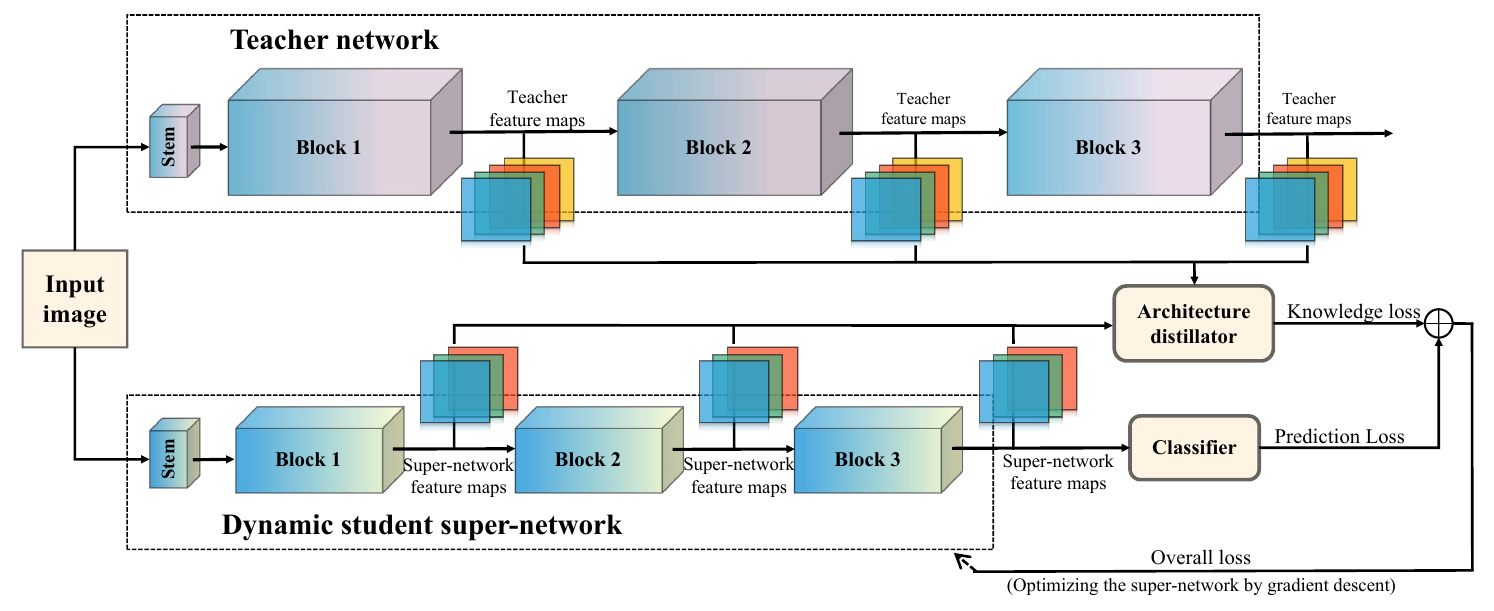}}
	\caption{A visual interpretation for DNAD. Both the teacher network and the student super-network are divided into $3$ blocks. In the forward process, both the teacher network and the student super-network receive the same input, and the intermediate feature maps of the teacher are used as another supervision signals, which aim to guide the optimization of the dynamic super-network and to regularize the one-level optimization of DARTS.}
\end{figure}

\section{Related Works}
\subsection{Model compression and knowledge distillation}
Model compression addresses the challenge of deploying large-scale NNs on resource-constrained devices by reducing model size and computational cost without significantly compromising performance. There are four typical approaches: network pruning \cite{wen2016learning,he2017channel,li2022dmpp}, parameter quantization \cite{rastegari2016xnor}, tensor decomposition \cite{yu2017compressing}, and KD \cite{hinton2015distilling,kim2018paraphrasing,heo2019comprehensive}.

Model compression reduces the size and computational costs of neural networks while preserving performance. Three primary approaches are widely adopted: network pruning removes redundant filters or neurons through structured sparsification \cite{he2017channel,li2022dmpp}; parameter quantization employs low-bit representations (e.g., 4-bit weights) to shrink memory footprint \cite{yang2019quantization,hubara2018quantized}; tensor decomposition leverages tensor ring/rank factorization to approximate high-dimensional parameters efficiently \cite{yin2021towards,wang2018wide}. Recent innovations like low-rank adaptation (LoRA) \cite{hu2021lora} further enable efficient finetuning of large models with minimal trainable parameters.

KD transfers knowledge from a well-trained large teacher model to a smaller student model. Response-based KD uses softed predicted probabilities from the teacher model as training signals \cite{hinton2015distilling,gou2021knowledge}. Recent advances include training targets adjusted by the teacher's prediction correctness \cite{meng2019conditional} and adaptive temperature coefficients \cite{li2023curriculum}. Feature-based KD leverages intermediate feature maps of the teacher to improve performance, with methods such as attention maps derived from feature maps \cite{DBLP:conf/iclr/ZagoruykoK17}, factor transfer for paraphrased information \cite{kim2018paraphrasing}, and cross-layer KD via attention allocation \cite{chen2021cross}. Relation-based KD utilizes knowledge from feature map relationships, such as Gram matrices (FSP) \cite{yim2017gift} and mutual information between hint layers \cite{passalis2020heterogeneous}.

\subsection{Neural architecture search}
NAS is expected to reduce the manual effort of designing neural architectures and has been advanced through various approaches, including reinforcement learning (RL) \cite{zoph2018learning}, evolutionary algorithms (EA) \cite{lu2020nsganetv2}, latent space optimization \cite{luo2018neural}, and Bayesian optimization \cite{kandasamy2018neural}. As an improved NAS approach, DARTS enhances the search efficiency by relaxing the discrete search space into a continuous one through an over-parameterized super-network \cite{liu2018darts}. Due to its computational efficiency, numerous follow-up works have further refined DARTS for better performance and efficiency \cite{xu2019pc, xie2018snas, bi2020gold, jing2023architecture, xue2023improved, chen2023mngnas, yang2024deeply}. For example, one study proposes an architecture regularizer to minimize the discretization discrepancy in DARTS \cite{jing2023architecture}, while another introduces progressive partial channel connections with channel attention (PA-DARTS) to reduce memory usage and operator competition \cite{xue2023improved}. Additionally, the deeply supervised block-wise NAS (DBNAS) method addresses the deep coupling issue in weight-sharing NAS methods \cite{yang2024deeply}.

Most NAS methods update architecture parameters \(\bm{\alpha}\) and network parameters \(\bm{w}\) using two distinct datasets (i.e., bi-level optimization), which is inefficient. DARTS attempted to optimize \(\bm{w}\) and \(\bm{\alpha}\) jointly on the same dataset (i.e., one-level optimization), but this approach yielded worse results than random search. Thus, regularizing one-level optimization in DARTS remains a critical challenge. GOLD-NAS \cite{bi2020gold} suggested that the failure of one-level optimization is due to overfitting of the super-network on small datasets, and proposed to use data augmentation techniques like auto-augment and cutout to mitigate this issue. However, our empirical ablation experiments show that the regularization provided by data augmentation is insufficient to improve the performance of one-level DARTS. The proposed DNAD approach in this paper aims to stabilize one-level DARTS by introducing additional supervision signals from a powerful teacher. Our experiments demonstrate that KD has a stronger regularization effect compared to data augmentation.

Several works have combined KD with NAS \cite{kang2020towards, li2020block, xie2022performance, DBLP:conf/icmcs/LiaoZWYLRYF22}. In \cite{li2020block}, KD is used to supervise block-wise architecture search and reduce representation shifts caused by weight-sharing. The primary role of KD here is to enhance the effectiveness of weight-sharing NAS by decomposing the large search space into smaller, block-wise spaces. A NAS-based KD framework \cite{DBLP:conf/icmcs/LiaoZWYLRYF22} aims to bridge the model capacity gap between the student network and ensemble models, focusing on improving KD rather than NAS. Additionally, MNGNAS introduces adaptive ensemble and perturbation-aware knowledge distillation to optimize feature map combination coefficients for better utilization of diverse architecture features \cite{chen2023mngnas}. PAMKD \cite{xie2022performance} addresses the collective failure problem in mutual knowledge distillation methods. KL-DNA \cite{singh2024kl} is a KD-based, latency-aware differentiable architecture search method for video motion magnification, considering computational complexity. Specifically, DistillDARTS also combines feature-based KD with DARTS \cite{DBLP:conf/icmcs/LiaoZWYLRYF22}. The key differences between DistillDARTS and our DNAD are: 1)DistillDARTS aims to alleviate the dominance of skip connections, while DNAD targets overfitting of the super-network through behavior imitation. 2) DistillDARTS is implemented within the constrained DARTS search space, whereas DNAD explores more flexible structures with knowledge regularization. 
\section{Super-network Progressive Shrinking}
This section presents the SNPS algorithm. The organizations are as follows. Part A gives a brief review on DARTS. Part B describes the elimination of topological constraints on cells. Then, Parts C-F provide the step-by-step descriptions on constructing the SNPS. 
\subsection{A brief overview of DARTS}
The neural cell is employed as the basic building component of neural architectures to avoid the direct searching of the whole network. In this paper, to improve the flexibility and the representation ability, neural architectures are searched in a search space where the same type of cells no longer share the same topology, which is different from previous DARTS-based approaches. 

From the perspective of information flow, a neural cell can be abstracted as a directed acyclic graph (DAG) of $N$ nodes, where the architecture is the only concern regardless of the countless inner parameters of operators. In the DAG, each node $x_{i}$ is a feature map and each directed edge $(i,j)$ is associated with a series of candidate operators $o^{(i,j)} \in \mathcal{O}$, where $\mathcal{O}$ is the operator space. Each cell has two input nodes $x_{0}$ and $x_{1}$, which are defined as the outputs of the former two cells, respectively. Each intermediate node is defined as the sum of information flows of all its predecessors, i.e., $x_{j} = \sum_{i<j}{o^{(i,j)}(x_{i})}$.
Then, the cell output is defined as the concatenation of all its intermediate nodes at channel dimension, i.e., $x_{N} = {\rm concat}(x_{2},x_{3},...,x_{N-1})$.

To make search space differentiable, $o^{(i,j)}(x_{i})$ is replaced by a differentiable mixture of all candidate operators in the operator space $\mathcal{O}$ as $\tilde{o}^{(i,j)}(x_{i}) = \sum_{o\in \mathcal{O}} {\sigma(\alpha_{o}^{i,j})o(x_{i})}$,
where $\alpha^{i,j}_{o} \in \mathbb{R} $ is the architecture parameter corresponding to the specific operator $o$ of edge $(i,j)$, and $\sigma(\cdot)$ is the architecture activation function (e.g., softmax, sigmoid and ReLU). 

After continuous relaxing of the search space, an over-parameterized super-network, whose edges are made up of the mixed operators, is constructed. The set of architecture parameters of all cells $\bm{\alpha} = \lbrace \alpha^{(i,j)} \rbrace$ is then referred to as the neural architecture, where the cell sequence number is omitted for convenience, and $\bm{w}$ is referred to as the network parameters associated with $\bm{\alpha}$. Let $\mathcal{L}_{\rm train}$ and $\mathcal{L}_{\rm val}$ be the loss functions on the training set and the validation set, respectively. Then, the goal of DARTS is to find an optimal architecture $\bm{\alpha}^{\ast}$ as
\begin{equation}
	\label{eq:nasgold}
	\begin{split}
		& \min_{\bm{\alpha}} {\mathcal{L}_{\rm val}(\bm{\alpha}^{\ast}, \bm{w}^{\ast})} \\
		& \mathrm{s.t.} \ \ {\bm{w}^{\ast}(\bm{\alpha}) = \arg\min_{\bm{w}}{\mathcal{L}_{\rm train}(\bm{\alpha}, \bm{w})} },
	\end{split}
\end{equation}
where $\bm{w}^{\ast}$ denotes the converged weights of the candidate architecture $\bm{\alpha}$.

Note that the equation \eqref{eq:nasgold} involves an inefficient bi-level optimization problem, where $\bm{\alpha}$ and $\bm{w}^{\ast}$ are the upper-level and lower-level variables, respectively. In order to approximate $\bm{w}^{\ast}(\bm{\alpha})$, DARTS employs a meta-learning-like strategy and tunes $\bm{w}$ with one-step spread out as $\bm{\tilde{w}}^{\ast} = \bm{w} - \varsigma \nabla_{\bm{w}}{\mathcal{L}_{\rm train}(\bm{\alpha},\bm{w})}$, where $\varsigma$ is the step size. Then, $\bm{\alpha}$ is optimized by $\nabla_{\bm{\alpha}}{\mathcal{L}_{\rm val}(\bm{\alpha},\bm{\tilde{w}}^{\ast})}$. However, due to the expensive second-order gradient calculation of $\nabla_{\bm{\alpha}}{\mathcal{L}_{\rm val}(\bm{\alpha},\bm{\tilde{w}}^{\ast})}$, most follow-up works adopted the faster first-order approximation \cite{chen2019progressive, xu2019pc}, i.e., $\bm{\alpha}$ is optimized by $\nabla_{\bm{\alpha}}{\mathcal{L}_{\rm val}(\bm{\alpha},\bm{w})}$ directly without $\bm{w}$ being replaced by $\bm{\tilde{w}}^{\ast}$.

\subsection{Enlarging the search space}
From the perspective of NAS, the over-parameterized super-network can be viewed as a DAG, where the upper structure of a neural network is the only concern and the inner parameters are ignored. In the DAG, edges represent the candidate operators, and nodes represent the intermediate feature maps. The importance degree of each edge can be measured by its architecture weights $\sigma{(\alpha_{o}^{i,j})}$, which are determined by corresponding architecture parameter $\alpha_{o}^{i,j}$. Note that, during the optimization of the super-network, some edges obtain small weights, and others obtain large weights. These weights can be viewed as the measure of the absolute contribution of these operators to the performance of the super-network, and this is the fundamental principle of DARTS-based approaches. 

It is worth pointing out that both the search space and the discretization mechanism of the super-network have significant impacts on the performance of searched neural architectures. 

For the search space, most current NAS approaches rely on a topology-sharing strategy, where the neural cells for constituting the architecture share the same topology. Specifically, in DARTS-based approaches, cells of the same type, namely normal cells or reduction cells, share the same topology, thus, the topology complexity of neural architectures is determined by two types of cells. Although it is reasonable to search architectures under the topology-constrained space in order to reduce the difficulty of NAS algorithms in exploring appropriate architectures and improving the transfer property of searched architectures, it is not reasonable enough if one expects to explore architectures with more competitive representation ability, since a neural network with a constrained architecture is not optimal. To the best of our knowledge, in previous DARTS-based approaches, GOLD-NAS is the only attempt to search neural architectures in an unconstrained search space, and the inspiration is inherited to design more potential powerful neural architectures. 

\subsection{Discussion on discretization mechanism}
An appropriate discretization mechanism derives a satisfactory discrete architecture from the dynamic super-network, but a poor one may push the super-network from optimal to local-optimal, or even an inappropriate location. Thus, the design of the discretization mechanism plays another important role for DARTS. For each intermediate node, most DARTS-based approaches retain the top-2 strongest non-zero operators collected from former nodes within the same cell. However, to acquire a higher accuracy, the super-network tends to keep the weights of all operators moderate. After all, each operator contributes more or less to the performance of the super-network regardless of the magnitude of the weights. However, this naive discretization mechanism may not evaluate the contribution of operators fairly especially when some operators have very close weights. Moreover, most DARTS-based approaches derive only one architecture in one search procedure, which is not convenient for one to pick up an architecture with a suitable computational burden according to the actual demands. To decrease the discretization gap between the discrete architecture and the super-network, $\rm DA^{2}S$ proposes to drive the weights of operators to be sparse under the regularization of an information entropy-like loss term \cite{tian2021discretization}. Nevertheless, $\rm DA^{2}S$ is still an approach based on the search space and discretization mechanism of DARTS, which means that it can only derive one architecture and the searched architectures are highly constrained by the manual rules.

An impartial discretization mechanism employed in this paper is to prune operators whose contributions are small enough to be negligible. For this purpose, the attention weight $\delta(\alpha_{k,j}^{o})$ with respect to the parameter (operator) $\alpha_{k,j}^{o}$ is calculated as 
\begin{equation}
	\delta(\alpha_{i,j}^{o}) = \frac{{\rm ReLU}({\alpha_{i,j}^{o}})} {\sum_{0 \le i'<j}\sum_{o'\in \mathcal{O}}{{\rm ReLU}({\alpha_{i',j}^{o'}})}},
\end{equation}
where $\rm ReLU(\cdot)$ is the rectified linear unit (ReLU). The attention weight $\delta(\alpha_{k,j}^{o})$ can also be designed as the softmax distribution of $\{ \alpha_{i,j}^{o}|0\le i< j, o \in \mathcal{O} \}$, but the empirical evidences imply that $\rm ReLU$ has a faster convergence speed than softmax. Note that $\delta(\alpha_{i,j}^{o})\ge0$ and $\sum_{i<j}\sum_{o\in\mathcal{O}}\delta(\alpha_{i,j}^{o})=1$, which means that there are competition relations among operators toward the same node. Then, to narrow the discrete gap, the operators of super-network will be pruned only when their attention weights are less than a predefined threshold (e.g., $0.01$). In this way, the super-network is shrunk from a dense architecture to a sparse one progressively and each intermediate super-network can be viewed as an appropriate successor of the initial super-network. 

\subsection{Performance-attentive sparsity entropy}
To guide the super-network toward a sparse structure while considering the representation performance, a sparsity entropy term is calculated for each intermediate node \( j \) as:
\begin{equation}
	\label{eq:sparsity_loss_node}
	\mathcal{L}^{j}_{\rm S}(\bm{\alpha}) = \sum_{i<j}\sum_{o\in\mathcal{O}}{{\rm log}(1 + K\delta(\alpha_{i,j}^{o}))},
\end{equation}
where \( K \) is a hyper-parameter controlling the flatness of \( \mathcal{L}^{j}_{\rm S}(\bm{\alpha}) \) and is set to 50 in this paper. The overall sparsity entropy for a neural cell is:
\begin{equation}
	\label{eq:sparsity_loss}
	\mathcal{L}_{\rm S}(\bm{\alpha}) = \sum_{1<j<M}{\mathcal{L}_{\rm S}^{j}}(\bm{\alpha}).
\end{equation}

In addition to the sparsity loss, the super-network minimizes the forward prediction loss:
\begin{equation}
	\label{eq:prediction_loss}
	\mathcal{L}_{\rm A}(\bm{\alpha}, \bm{w}) = \mathbb{E}_{(x,y^{\star})\in \mathcal{D}_{\rm train}} [ {\rm CE}(f(x),y^{\star}) ],
\end{equation}
where \( {\rm CE}(f(x),y^{\star}) \) is the cross entropy between the network mapping \( f(x) \) and the target \( y^{\star} \).

When only the sparsity loss \( \mathcal{L}_{\rm S}(\bm{\alpha}) \) is considered, the weights of all operators will be optimized toward an approximate one-hot distribution. To balance the model performance and sparsity, the loss function is designed as:
\begin{equation}
	\label{eq:fake_overall_loss}
	\mathcal{L}(\bm{\alpha}, \bm{w}) = \mathcal{L}_{\rm A}(\bm{\alpha}, \bm{w}) + \gamma\mathcal{L}_{\rm S}(\bm{\alpha}),
\end{equation}
where \( \gamma \) is a coefficient for controlling the tradeoff between different objectives.

However, optimizing \( \mathcal{L}_{\rm A}(\bm{\alpha}, \bm{w}) \) is more challenging than \( \mathcal{L}_{\rm S}(\bm{\alpha}) \) because it is correlated with the model performance and requires more time to optimize and evaluate. Therefore, \( \gamma \) must be carefully adjusted to avoid either excessive focus on sparsity or insufficient sparsity. To this end, a performance-attentive loss function is designed as
\begin{equation}
	\mathcal{L}(\bm{\alpha},\bm{w}) = \mathcal{L}_{\rm A}(\bm{\alpha},\bm{w})\cdot(1 + \gamma \mathcal{L}_{\rm S}(\bm{\alpha})) + \gamma \mu \cdot \mathcal{L}_{\rm S}(\bm{\alpha}),
	\label{eq:overall_loss}
\end{equation}
where \( \gamma \) is the multiplicative sparsity coefficient, and \( \mu \) is the additive sparsity coefficient. The initial value \( \mu_0 \) of \( \mu \) is set to be much smaller than 1, ensuring that \( \mu_0 \cdot \gamma_0 \ll \gamma_0 \) and \( \mathcal{L} \approx \mathcal{L}_{\rm A}\cdot(1 + \gamma \mathcal{L}_{\rm S}) \) at the early stage. The gradient of the sparsity entropy with respect to architecture \( \bm{\alpha} \) is
\begin{align}
	\nabla_{\bm{\alpha}}(\gamma \mathcal{L}_{\rm A}\cdot\mathcal{L}_{\rm S}) = \gamma \mathcal{L}_{\rm S}\cdot\nabla_{\bm{\alpha}}(\mathcal{L}_{\rm A}) + \gamma \mathcal{L}_{\rm A}\cdot\nabla_{\bm{\alpha}}(\mathcal{L}_{\rm S}).
\end{align}
If \( \mathcal{L}_{\rm A}\cdot\mathcal{L}_{\rm S} \) is the only concern, the system may reach an equilibrium state where neither \( \mathcal{L}_{\rm A} \) nor \( \mathcal{L}_{\rm S} \) decreases further. The term \( \gamma \mu \cdot \mathcal{L}_{\rm S} \) ensures continuous sparsity by applying unconditional pressure on \( \mathcal{L}_{\rm S} \). An upper bound \( \gamma_{\rm max} \) is set for \( \gamma \), and if \( \gamma \) frequently reaches this bound, \( \mu \) is increased to further encourage sparsity.
 
\subsection{The progressive shrinking of super-network}
The overall procedure of the super-network progressive shrinking (SNPS) algorithm is summarized in Algorithm \ref{alg:super_network_shrinking}. During the optimization, the dynamic super-network is shrunk from a dense architecture to a sparse one by pruning less important operators. For each training step, an expected number \( n_{\rm expect} \) of pruned operators is set, and the actual number \( n_{\rm prune} \) is compared to this expectation. If \( n_{\rm prune} < n_{\rm expect} \), the sparsity entropy is increased by raising \( \gamma \) (or \( \mu \) if \( \gamma \) has reached \( \gamma_{\rm max} \)). Otherwise, \( \gamma \) is reduced. The actual number of pruned operators is smoothed using an exponential moving average \( \overline{n}_{\rm prune} = (1 - \rho)n_{\rm prune} + \rho\overline{n}_{\rm prune}' \), where \( \rho \) is the smooth coefficient and \( \overline{n}_{\rm prune}' \) is the previous smoothed value.

\subsection{Convergence Analysis of the Dynamic Sparsity Control}

The convergence of the sparsity entropy \( \mathcal{L}_{\rm S}(\bm{\alpha}) \) in the SNPS algorithm is guaranteed through the collaborative adaptation of the multiplicative coefficient \( \gamma \) and additive coefficient \( \mu \). The roles of \( \gamma \) and \( \mu \) in ensuring the convergence of \( \mathcal{L}_{\rm S} \) toward \( \mathcal{L}_{\rm min} \) are analyzed through three perspectives.
\paragraph{Closed-loop control of $\gamma$ for steady pruning}
As a closed-loop controller, the multiplicative coefficient \( \gamma \) dynamically adjusts the intensity of sparsity regularization based on the discrepancy between the actual and expected pruning rates. The gradient of overall loss with respect to architecture \( \bm{\alpha} \) is
\begin{align}
	\label{eq:gradient}
	\nabla_{\bm{\alpha}} \mathcal{L}\left( \bm{\alpha}, \bm{w} \right ) = \underbrace{\left( 1 + \gamma\mathcal{L}_{\rm S} \right)\nabla_{\bm{\alpha}}\mathcal{L}_{\rm A}}_{\text{Performance term}} +  \underbrace{\left( \gamma \mathcal{L}_{\rm A} +  \gamma \mu \right) \nabla_{\bm{\alpha}}\mathcal{L}_{\rm S}}_{\text{Sparsity term}}.
\end{align}
When \( \overline{n}_{\rm prune} < n_{\rm expect} \), \( \gamma \) is amplified by a factor of \( \varphi > 1 \). Conversely, if \( \overline{n}_{\rm prune} \geq n_{\rm expect} \), \( \gamma \) is attenuated by a factor of \( k < 1 \) to prioritize the model performance. This feedback mechanism ensures that \( \overline{n}_{\rm prune} \) oscillates around \( n_{\rm expect} \). The boundedness of \( \gamma \) within \( [\gamma_0, \gamma_{\rm max}] \) prevents excessive perturbations and stabilizes the optimization trajectory.

\paragraph{Enlarging $\mu$ to break the gradient equilibrium} 
When $\gamma$ reaches its upper bound $\gamma_{\rm max}$ consistently, $\nabla_{\bm{\alpha}} \mathcal{L}\left( \bm{\alpha}, \bm{w} \right )$ has reached the gradient equilibrium as
\begin{align}
	\left( 1 + \gamma \mathcal{L}_{\rm S} \right)\nabla_{\bm{\alpha}}\mathcal{L}_{\rm A} + \left( \gamma \mathcal{L}_{\rm A} +  \gamma \mu \right) \nabla_{\bm{\alpha}}\mathcal{L}_{\rm S} \approx 0.
\end{align}
At this moment, the current $\gamma$ and $\mu$ cannot regulate the $\overline{n}_{\rm prune}$ to satisfy $n_{\rm expect}$. Then, the incremental updating of $\mu$ by $\phi > 1$ amplifies the weight of $\mathcal{L}_{\rm S}$ in the overall loss, which makes the enlarged residual term $\mu \nabla \mathcal{L}_{\rm S}$ disrupt the gradient equilibrium such that $\Vert  \left( \gamma \mathcal{L}_{\rm A} +  \gamma \mu \right) \nabla_{\bm{\alpha}}\mathcal{L}_{\rm S}  \Vert > \Vert 	\left( 1 + \gamma \mathcal{L}_{\rm S} \right)\nabla_{\bm{\alpha}}\mathcal{L}_{\rm A} \Vert$, where the gradient norm of sparsity term is larger than the gradient norm of performance term. Furthermore, since $\gamma \mu \nabla_{\bm{\alpha}}\mathcal{L}_{\rm S}$ is independent of $\mathcal{L}_{A}$, whenever the $\nabla_{\bm{\alpha}} \mathcal{L}\left( \bm{\alpha}, \bm{w} \right )$ reaches the gradient equilibrium, the adjustment of $\mu$ will reintroduce a gradient component that prioritizes the sparsity over the model performance. 

\paragraph{Convergence guarantee} Based on the above analysis, the joint adaptation of \( \gamma \) and \( \mu \) guarantees that \( \mathcal{L}_{\rm S}(\bm{\alpha}) \) monotonically decreases until reaching the predefined minimum entropy \( \mathcal{L}_{\rm S}^{\rm min} \). Suppose $\mathcal{L}_{\rm S}(\bm{\alpha}) > \mathcal{L}_{\rm S}^{\rm min}$, the following conclusions hold.
\begin{itemize}
	\item The multiplicative term $\gamma \mathcal{L}_{\rm A}  \mathcal{L}_{\rm S}$ causes $\mathcal{L}_{\rm S}$ to decrease in proportion to the model performance $\mathcal{L}_{\rm A}$, while the additive term $\gamma \mu  \mathcal{L}_{\rm S}$ applies unconditional pressure on $\mathcal{L}_{\rm S}$.
	\item When $\gamma$ reaches $\gamma_{\rm max}$ consistently, $\mu$ grows exponentially by $\mu \leftarrow \phi \mu$ to counteract the gradient equilibrium, such that the sparsity entropy can be further reduced.
	\item Once $\mathcal{L}_{\rm S}(\bm{\alpha})$ approaches $\mathcal{L}_{\rm S}^{\rm min}$, the pruning process is terminated. 
\end{itemize}
\begin{algorithm}[h]
	\caption{Super-network progressive shrinking (SNPS)}
	\label{alg:super_network_shrinking}
	\begin{algorithmic}[1]
		\STATE{\textbf{Initialization}: The training set $\mathcal{D}_{\rm train}$; the dynamic super-network, which is characterized by architecture parameters $\bm \alpha$ and network parameters $\bm w$;  the multiplicative sparsity coefficient $\gamma \leftarrow \gamma_{0}$; the expansion rate $\varphi$ and attenuation rate $k$ for $\gamma$, where $\varphi > 1$ and $0 < k< 1$; the maximum value $\gamma_{\rm max}$ of $\gamma$; the maximum step number $\tau_{\rm max}$ for $\gamma > \gamma_{\rm max}$; the additive sparsity coefficient $\mu \leftarrow \mu_{0}$ (where $\mu_{0} \ll 1$); the expansion rate $\phi$ for $\mu$, where $\phi > 1$; the expected number $n_{\rm expect}$ of pruned operators for each step; the moving average coefficient $\rho$; the discrete threshold $\epsilon$ for an operator; the $\it Interval$ for discretizing the super-network; the accumulated $step \leftarrow 0$ within an interval; the discrete architecture set $\mathcal{A}$; the minimum sparsity entropy $\mathcal{L}_{\rm S}^{\rm min}$.}
		\WHILE{$\mathcal{L}_{\rm S}(\alpha) > \mathcal{L}_{\rm S}^{\rm min}$}
		\STATE{Update $step \leftarrow step + 1$;}
		\STATE{Sample a mini-batch training data $(X,Y)$ from $\mathcal{D}_{\rm train}$ randomly;}		
		\STATE{Update $\bm \alpha$ and $\bm w$ simultaneously by minimizing the loss \eqref{eq:overall_loss} on $(X,Y)$ with gradient-descent algorithm; }
		\STATE{Prune the operators of super-network whose attention weights are smaller than $\epsilon$;}
		\STATE{Obtain the actual prune number $n_{\rm prune}$;}
		\STATE{Smooth $n_{\rm prune}$ by $\overline{n}_{\rm prune} \leftarrow (1 - \rho)n_{\rm prune} + \rho\overline{n}_{\rm prune}$;}
		\STATE{\textbf{if} $\overline{n}_{\rm prune} \ge n_{\rm expect}$, \textbf{then}, $\gamma \leftarrow k\gamma $ ($0<k<1$) \textbf{else} $\gamma \leftarrow \varphi\gamma $ ($\varphi > 1$);}
		\STATE{\textbf{if} $\gamma > \gamma_{\rm max}$, \textbf{then}, $\tau \leftarrow \tau + 1$ \textbf{else} $\tau \leftarrow 0$;}
		\STATE{\textbf{if} $\tau > \tau_{\rm max}$, \textbf{then}, $\mu \leftarrow \phi \mu $ ($\phi > 1$) and $\tau \leftarrow 0$;}
		\STATE{\textbf{if} $step > {\it Interval}$, \textbf{then}, append the pruned architecture of super-network to $\mathcal{A}$ and $step \leftarrow 0$.}
		\ENDWHILE
		\STATE{\textbf{Return}: $\mathcal{A}$}
	\end{algorithmic}	
\end{algorithm}

\section{Neural Architecture Distillation}

This section presents the DNAD algorithm. Parts A and B introduce the KD method employed in DNAD and describe the calculation of the knowledge loss. Part C combines SNPS with KD to formalize the DNAD algorithm. Part D analyzes the role of KD in SNPS. Finally, Part E introduces other types of KD methods which can also be used in SNPS. 
\subsection{Activation-based attention}
The knowledge of a teacher can be represented by different forms, such as the softmax output \cite{hinton2015distilling} and intermediate attention maps \cite{romero2014fitnets}. As a feature-based distillation (FD) method, the attention transfer (AT) algorithm \cite{DBLP:conf/iclr/ZagoruykoK17}, which presents advantages to other KD methods, is employed in this paper. 

Considering an intermediate layer of a convolutional neural network and its 3D feature map $A\in \mathbb{R}^{C\times H \times W}$, where $C$ is the channel dimension and $H\times W$ is the spatial dimensions, the activation-based mapping function $\mathcal{F} \colon \mathbb{R}^{C\times H \times W } \rightarrow \mathbb{R}^{H \times W}$ with respect to this layer takes the 3D tensor $A$ as inputs and outputs a 2D spatial attention map, as 
\begin{equation}
	\mathcal{F}(A)=\sum_{i=1}^{C}|A_{i}|,
\end{equation} 
where $A_{i}\in \mathbb{R}^{H \times W}$ is the 2D feature map of $A$ in the $i$th channel. The activation-based function $\mathcal{F}$ makes it possible to calculate the similarity loss between feature maps with different channel dimensions as long as their spatial resolutions are compatible. 

\subsection{Calculation of knowledge loss}
To implement AT, both the teacher network and the student super-network are required to be divided into $M$ blocks by different depths. Blocks of the same relative depth are required to have the same spatial resolution for similarity calculation. Given an input data $x$, let $f_{S}(x)$ and $f_{T}(x)$ be the mappings of the super-network and the teacher network, respectively. Since the architecture has been divided into $M$ blocks, either of $f_{S}(x)$ or $f_{ T}(x)$ can be viewed as the composition of $M$ functions. Without loss of generality, the output of the $k$th block of the super-network is denoted by $f_{S}^{k}(x)= (g_{1}\circ\cdots\circ g_{k})(x)$, where $g_{i}(\cdot)$ denotes the mapping of $i$th block. To calculate the knowledge loss, we need to collect the feature maps of the super-network and teacher network. Mathematically, given the input $x$, we collect feature maps $A_{S}^{k} = f_{S}^{k}(x)$ and $A_{T}^{k}=f_{T}^{k}(x)$ and calculate the knowledge loss by
\begin{equation}
	\label{eq:knowledge_loss}
	\mathcal{L}_{\rm AT}(\bm{\alpha}, \bm{w}) = \mathbb{E}_{(x,y^{\star})\in \mathcal{D}_{\rm train}} \Bigg[\sum_{k=1}^{M} \Bigg{\Vert}  \frac{Q_{S}^{k}}{\Vert Q_{S}^{k} \Vert} - \frac{Q_{T}^{k}}{\Vert Q_{T}^{k} \Vert} \Bigg{\Vert}^{2} \Bigg],
\end{equation}
where $\Vert \cdot \Vert$ is the $L_{2}$ norm, and both $Q_{S}^{k} = \mathcal{F}(A_{S}^{k})$ and  $Q_{T}^{k} = \mathcal{F}(A_{T}^{k})$ are activation-based attention maps.
Then, by adding the forward prediction loss, the knowledge loss of the DNAD super-network is calculated by 
\begin{equation}
	\label{eq:loss}
	\mathcal{L}_{\rm KD}(\bm{\alpha}, \bm{w}) = \mathcal{L}_{\rm A}(\bm{\alpha}, \bm{w}) + \frac{\beta}{2} \mathcal{L}_{\rm AT}(\bm{\alpha}, \bm{w}),
\end{equation}
where $\beta$ is the balance coefficient chosen as $10^{3}$ following the basic setting \cite{DBLP:conf/iclr/ZagoruykoK17}. It is worth pointing out that the block-wise strategy is reasonable since each dynamic block of the super-network can be optimized to match the corresponding block-wise behavior of the teacher. In this way, the super-network absorbs knowledge more properly as the teacher has delivered more behavior information by block-wise strategy.

\subsection{Differentiable neural architecture distillation}
To minimize the knowledge loss \eqref{eq:loss}, the super-network is required to adjust its mapping to match teacher's mapping, i.e., the \emph{search by imitating}, by optimizing architecture and network parameters using a stochastic gradient optimizer such as SGD and Adam. To combine the proposed SNPS method with the KD technique, \eqref{eq:overall_loss} is replaced with \eqref{eq:loss}. Consequently, we obtain the overall loss for the DNAD super-network as
\begin{equation}
	\mathcal{L}(\bm{\alpha},\bm{w}) = \mathcal{L}_{\rm KD}(\bm{\alpha},\bm{w})(1 + \gamma \mathcal{L}_{\rm S}(\bm{\alpha})) + \gamma \mu  \mathcal{L}_{\rm S}(\bm{\alpha}).
	\label{eq:overall_loss_2}
\end{equation}

The developed DNAD is summarized in Algorithm \ref{alg:one_training_step}.
\begin{algorithm}[h]
	\caption{Differentiable neural architecture distillation}
	\label{alg:one_training_step}
	\begin{algorithmic}[1]
		\STATE{\textbf{Input}: the SNPS algorithm, a well-trained teacher network for distilling knowledge.}
		\STATE{Replace the loss function \eqref{eq:overall_loss} of dynamic super-network by knowledge loss \eqref{eq:loss}, i.e., $\mathcal{L}(\bm{\alpha},\bm{w}) = \mathcal{L}_{\rm KD}(\bm{\alpha},\bm{w})(1 + \gamma \mathcal{L}_{\rm S}(\bm{\alpha})) + \gamma \mu  \mathcal{L}_{\rm S}(\bm{\alpha})$;} \\
		\STATE{Obtain the architecture set $\mathcal{A}$ by SNPS (Algorithm 1).
		\STATE{\textbf{Output:} $\mathcal{A}$ }}
	\end{algorithmic}
\end{algorithm}

\subsection{Discussion on other distillation methods}
The implementation of KD means that the feature maps of the teacher and student are collected. However, in this relatively restrictive approach, the teacher is required to own an analogous structure with the student in order to calculate the knowledge loss. As another distillation method, response-based distillation, is developed based on the response of the last output layer of neural networks. The most popular response-based distillation method uses the soft target (ST) \cite{hinton2015distilling}. In particular, ST refers to the predictive probability that an input belongs to a specific class, and is calculated by
\begin{equation}
	p(z_{i}, t) = \frac{{\rm exp}(z_{i}/t)}{\sum_{j}{\rm exp}(z_{j}/t)},
\end{equation}
where $z_{i}$ is the logit for the $i$th class, and $t$ is the temperature factor for controlling the smoothness of softmax function. Following the suggestion in \cite{hinton2015distilling}, we select $t=4$ in this paper. Then, the response-based knowledge loss is calculated by 
\begin{equation}
	\mathcal{L}_{\rm{ST}}(\bm{\alpha}, \bm{w}) = D_{\rm {KL}} \big( p(\bm{z}_{\rm S})||p(\bm{z}_{\rm T}) \big),
\end{equation}
where $p(\bm{z}_{\rm S}) = [...,p(z_{i}^{\rm {S}},t),...]$ and $p(\bm{z}_{\rm T}) = [...,p(z_{i}^{\rm {T}},t),...]$ are discrete probability distributions of student and teacher networks, respectively, and $D_{\rm {KL}}(\cdot||\cdot)$ is the Kullback-Leibler divergence between two distributions. 

If the ST knowledge is employed, we obtain another version of DNAD algorithm, whose knowledge loss is calculated by
\begin{equation}
	\label{eq:st_loss}
	\mathcal{L}_{\rm{KD}}(\bm{\alpha}, \bm{w}) = \mathcal{L}_{\rm A}(\bm{\alpha}, \bm{w}) + t^{2}\mathcal{L}_{\rm ST}(\bm{\alpha}, \bm{w}).
\end{equation}

Furthermore, if both the AT and ST are employed, the third version of DNAD algorithm is obtained with knowledge loss being calculated by
\begin{equation}
	\label{eq:sa_loss}
	\mathcal{L}_{\rm{KD}}(\bm{\alpha}, \bm{w}) = \mathcal{L}_{\rm A}(\bm{\alpha}, \bm{w}) +\frac{\beta}{2}\mathcal{L}_{\rm{AT}}(\bm{\alpha}, \bm{w}) + t^{2}\mathcal{L}_{\rm ST}(\bm{\alpha}, \bm{w}).
\end{equation}
In the ablation experiments, all three types of knowledge loss, i.e., \eqref{eq:loss}, \eqref{eq:st_loss} and \eqref{eq:sa_loss}, are tested. However, the results show that only the AT loss \eqref{eq:loss} and ST + AT loss \eqref{eq:sa_loss} improve the performance of SNPS. In this paper, we refer the knowledge loss to the AT loss \eqref{eq:loss} without a specific emphasis.

\subsection{Analysis on the role of feature-based KD}
By decomposing complex tasks into hierarchical subtasks through the attention map mimicry, the student super-network not only resolves the difficulty of direct end-to-end optimization, but also implicitly encodes the teacher’s generalization strategies. This hierarchical alignment mitigates shortcut learning behaviors, such as the skip connection dominance in DARTS by enforcing structured feature interactions. For instance, AT explicitly guides the student to replicate the teacher’s spatial attention patterns across layers, which ensures granular feature representation learning. Unlike the ST distillation which focuses solely on probabilistic outputs, AT provides intermediate supervision that enriches gradient propagation dynamics. These gradients counteract vanishing issues in deeper architectures while reducing over-reliance on parameter-free operators (e.g., pooling layers), which is a critical weakness in unstable search spaces.  

Furthermore, the teacher’s behavioral imitation serves as a dual-purpose mechanism, i.e., it minimizes the empirical risk on training data while transferring implicit regularization patterns to enhance generalization. When datasets are limited, this approach acts as a safeguard against over-fitting by aligning the student’s decision-making logic with the teacher’s robust internal semantics. For example, intermediate supervision in AT forces the student to learn semantically meaningful feature hierarchies, which avoids superficial shortcut solutions that ST-based methods might tolerate. The ablation results (e.g., DNAD-AN-C achieving 3.95\% error vs. 4.10\% for ST variants) empirically validate that the hierarchical knowledge transfer strengthens the architectural stability and performance. Thus, by unifying the task decomposition, gradient refinement, and generalization-oriented behavioral alignment, the AT distillation brings more benefits to NAS than ST distillation.

\section{Experimental Results on CIFAR-10}
In this section, we evaluate the performance of the proposed SNPS and DNAD algorithms on the CIFAR-10 dataset.
\subsection{Hyper-parameter settings}
\subsubsection{Super-network}
The super-network consists of 14 neural cells in both search and re-train procedures. The initial channel number is set to be $20$. By convention, two reduction cells are located at the $1/3$ and $2/3$ network depths, respectively. The network parameters $\bm w$ is optimized by an adaptive moment estimation (Adam) optimizer with a fixed learning rate of $5\times10^{-3}$, an adaptive momentum $\in (0.5, 0.999)$ and a weight decay of $3\times10^{-4}$. The architecture $\bm \alpha$ is optimized by a stochastic gradient descent (SGD) optimizer with a learning rate of $0.2$ and a momentum of $0.9$. The weight decay is set to be $3\times10^{-4}$. To warm up the super-network, it is trained for five epochs without any constraints. The batch size is set to be 96. Cutout is added as the additional data-augment method with 16$\times$16 size. The operator space of super-network contains three types of operators, which are skip connection, separable convolution $3\times3$, and dilated convolution $5\times5$. 

\subsubsection{SNPS} The initial multiplicative sparsity coefficient $\gamma_{0}$ is $1\times 10 ^{-4}$. The corresponding expansion rate $\varphi$ is $1.01$ and attenuation rate $k$ is $0.99$. The settings of $\varphi$ and $k$ are unrestricted relatively as long as $\varphi > 1$ and $0< k < 1$. The maximum step-number $\tau_{\rm max}$ is $50$. The upper bound $\gamma_{\rm max}$ is set to be $0.01$. The initial additive sparsity coefficient $\mu_{0}$ is $0.01$. The expansion rate $\phi$ for $\mu$ is $1.5$. The discrete threshold $\epsilon$ is set to be $0.01$. The operators of super-network will be pruned only when their attention weights are less than $\epsilon$. The moving average coefficient $\rho$ is $0.90$, which is used to reduce the fluctuation of $n_{\rm prune}$. The expected prune-number $n_{\rm expect}$ is $0.003$. When other parameters remain unchanged, a larger $n_{\rm expect}$ leads to a faster search procedure. The discrete $\it Interval$ is set to be the number of mini-batch in an epoch. The minimum sparsity entropy $\mathcal{L}_{S}^{\rm min}$ is $200$.
\subsubsection{DNAD} For DNAD, an architecture searched by SNPS is used as the teacher network, which achieves the error rate of $2.60\%$ on CIFAR-10. In particular, the teacher network remains unchanged unless a specific statement is presented. By convention of DARTS-based methods, there are three different resolutions in the super-network. Thus, both the teacher network and student super-network are divided into $M=3$ blocks. In this section, the feature-based distillation loss \eqref{eq:knowledge_loss} is employed. The distillation positions are located at the end of each block, which is inspired by \cite{heo2019comprehensive}. 
\subsubsection{Discrete architectures} The channel number is set to be 36. To reduce the influence of randomness, each architecture is re-trained for three times, and the average of highest error rates is reported. By convention, each architecture is trained for 600 epochs with a batch size of 96. The network parameters are optimized by an SGD optimizer with an initial learning rate of $0.025$ and a momentum of $0.9$. The learning rate is annealed to zero following a cosine schedule. The weight decay is set to be $3\times10^{-4}$. In addition, cutout with $16\times16$ size and path dropout with a probability of $0.3$, and an auxiliary tower with a weight of $0.4$ are employed. 
\subsection{Evaluation results on CIFAR-10}
\begin{figure}[htbp]	
	\centering{\includegraphics[width=0.50\textwidth]{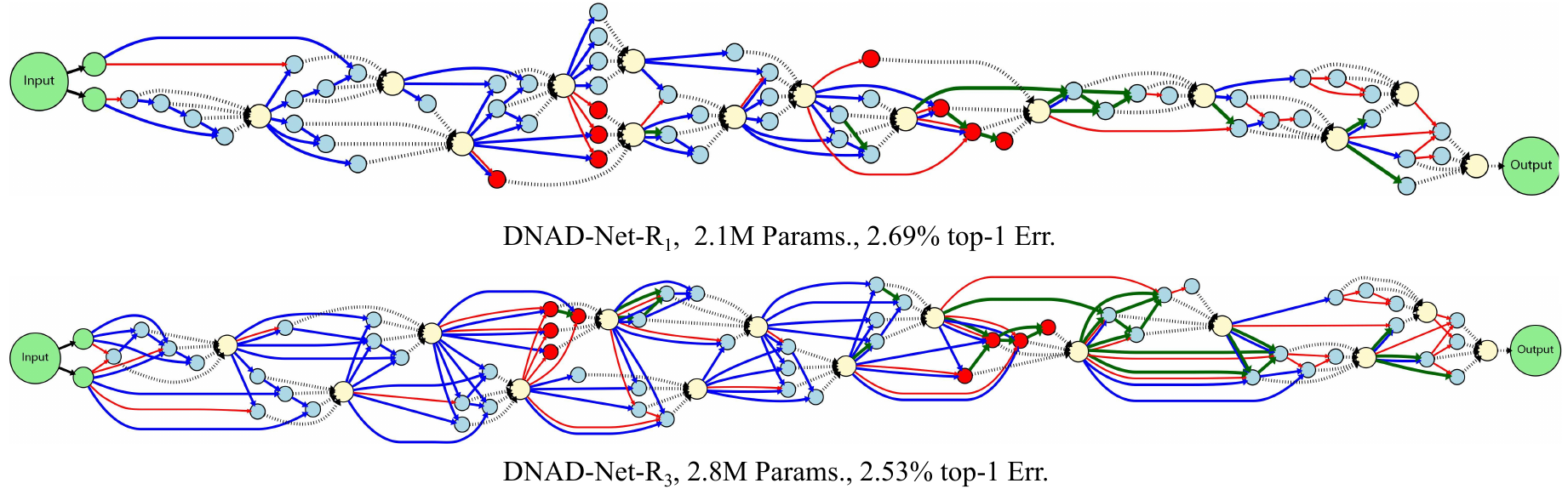}}
	\caption{Two neural architectures searched by DNAD on CIFAR-10. In this figure, neural architectures are named in an increasing sequence of model size. Nodes with different colors denote the feature maps of neural networks. In particular, green nodes denote the inputs and the outputs. Blue nodes denote the intermediate nodes of normal cells. Red nodes denote the intermediate nodes of reduction cells. Cream yellow nodes denote the output nodes of both normal cells and reduction cells. There are no extra nodes with different colors to denote the input nodes of cells, since the inputs of each cell are actually the outputs of its former two cells. Solid lines with different colors denote different types of operators. Skip-connect is marked in red line, sep-conv-3 is marked in blue line and dil-conv-5 is marked in green line. The gray dotted lines denote the concatenation of intermediate nodes in the channel dimension.}
	\label{fig:NADD_nets}
\end{figure}
\begin{table}[htbp]
	\centering
	\caption{Evaluation results of recent state-of-the-art NAS approaches on CIFAR-10}
	\label{tab:cifar10_results}
	\begin{tabular}{l|c|c|c}
		\toprule
		\multirow{2}{*}{{Architecture}} & {Params} & {Test Err.} & {Search cost} \\
		{} & {(M)} & ($\bm{\%}$) & {(Gpu-days)} \\
		\midrule
		NASNet-A \cite{zoph2018learning} & 3.3 & 2.65 & 1800 \\
		DARTS(1st) \cite{liu2018darts} & 3.3 & 3.00 $\pm$ 0.14 & 1 \\
		DARTS(2nd) \cite{liu2018darts} & 3.3 & 2.76 $\pm$ 0.09 & 4 \\
		SNAS(moderate) \cite{xie2018snas} & 2.8 & 2.85 $\pm$ 0.02 & 1.5 \\
		P-DARTS \cite{chen2019progressive} & 3.4 & 2.50 & 0.3  \\
		PC-DARTS \cite{xu2019pc} & 3.6 & 2.57 & 0.1 \\ 
		CARS-I \cite{yang2020cars} & 3.6 & 2.62 & 0.4 \\
		GDAS(FRC) \cite{dong2019searching} & 2.5 & 2.82 & 0.4 \\ 		
		DA$^{2}$S \cite{tian2021discretization} & 3.4 & 2.51 $\pm$ 0.09 & 0.3 \\
		SDARTS-ADV \cite{chen2020stabilizing} & 3.3 & 2.61 $\pm$ 0.02 & 1.3 \\
		DOTS \cite{gu2021dots} & 3.5 & 2.49 $\pm$ 0.06 & 0.3 \\
		CDARTS \cite{9720178} & 3.9 & 2.48 $\pm$ 0.04 & 0.3 \\
		MiLeNAS \cite{he2020milenas} & 2.9 & 2.80 $\pm$ 0.04 & 0.3 \\ 
		LFR-DARTS \cite{DBLP:journals/soco/HaoZ22} & 2.7 & 2.65 & 0.5 \\
		DistillDARTS-SD(1st) \cite{DBLP:conf/icmcs/LiaoZWYLRYF22} & 4.0 & 2.64 $\pm$ 0.02 & 0.5 \\
		DistillDARTS-SD(2nd) \cite{DBLP:conf/icmcs/LiaoZWYLRYF22} & 4.2 & 2.66 $\pm$ 0.08 & 2.0 \\
		DistillDARTS-KD(1st) \cite{DBLP:conf/icmcs/LiaoZWYLRYF22} & 3.7 & 2.59 $\pm$ 0.05 & 0.5 \\
		DistillDARTS-KD(2nd) \cite{DBLP:conf/icmcs/LiaoZWYLRYF22} & 3.5 & 2.48 $\pm$ 0.05 & 2.3 \\
		$\beta$-DARTS \cite{DBLP:conf/cvpr/YeL00FO22} & 3.8 & 2.51 $\pm$ 0.07 & 0.4 \\
		EoiNAS \cite{9432795} & 3.4 & 2.50 & 0.6 \\
		ReCNAS(w/o SE) \cite{9836970} & 4.1 & 2.52 $\pm$ 0.09 & 0.3 \\  
		\midrule
		Gold-NAS-G \cite{bi2020gold} & 2.2 & 2.75 $\pm$ 0.09 & \multirow{3}{*}{1.1} \\
		Gold-NAS-H \cite{bi2020gold} & 2.5 & 2.70 $\pm$ 0.07 & {} \\
		Gold-NAS-I \cite{bi2020gold} & 2.9 & 2.61 $\pm$ 0.10 & {} \\
		\midrule
		{SNPS-Net-R$_{1}$} & 2.2 & 2.73 $\pm$ 0.05 & \multirow{4}{*}{0.5} \\
		{SNPS-Net-R$_{2}$} & 2.5 & 2.61 $\pm$ 0.12 & {} \\
		{SNPS-Net-R$_{3}$} & 2.9 & 2.57 $\pm$ 0.13 & {} \\
		{SNPS-Net-R$_{4}$} & 3.5 & 2.53 $\pm$ 0.06 & {} \\
		\midrule
		{DNAD-Net-R$_{1}$} & 2.1 & 2.69 $\pm$ 0.05 & \multirow{4}{*}{0.6} \\
		{DNAD-Net-R$_{2}$} & 2.5 & 2.59 $\pm$ 0.09 & {} \\
		{DNAD-Net-R$_{3}$} & 2.8 & 2.53 $\pm$ 0.08 & {} \\
		{DNAD-Net-R$_{4}$} & 3.3 & 2.47 $\pm$ 0.04 & {} \\
		\bottomrule
	\end{tabular}
\end{table}
\begin{figure}[htbp]
	\centerline{\includegraphics[width=0.30\textwidth]{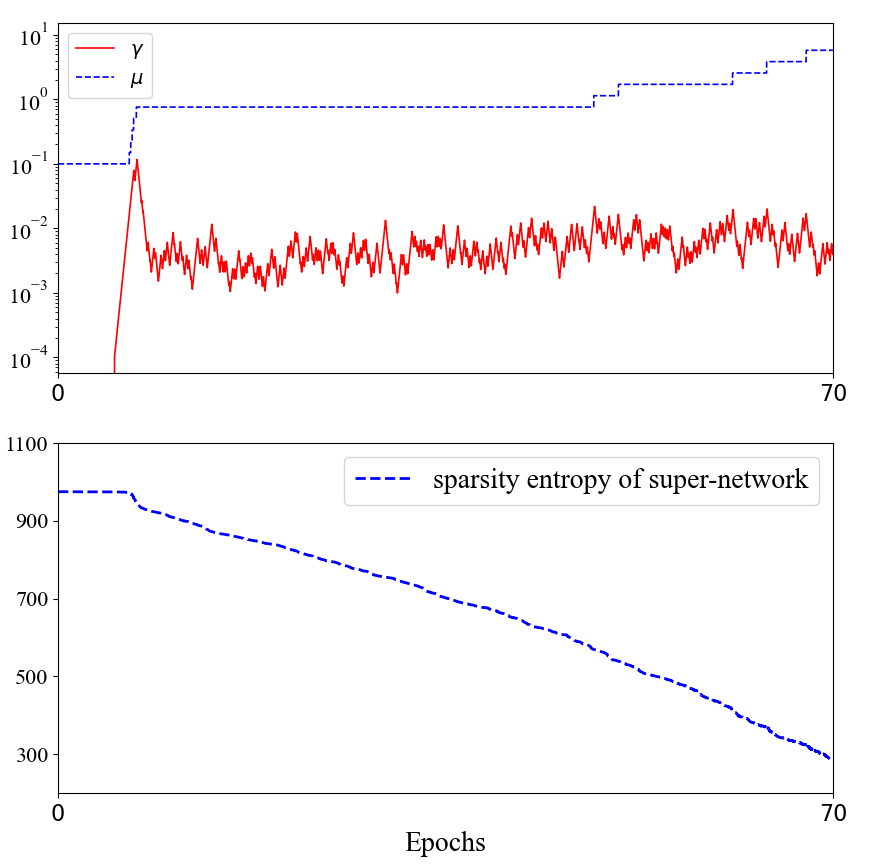}}
	\caption{The intermediate search procedure of DNAD. The top sub-figure presents the evolutions of multiplicative sparsity coefficient $\gamma$ and the additive sparsity coefficient $\mu$. The bottom sub-figure presents the evolution of sparsity entropy of the super-network during the search procedure.}
	\label{fig:search_procedure}
\end{figure}
\begin{figure*}[htbp]
	\centerline{\includegraphics[width=0.80\textwidth]{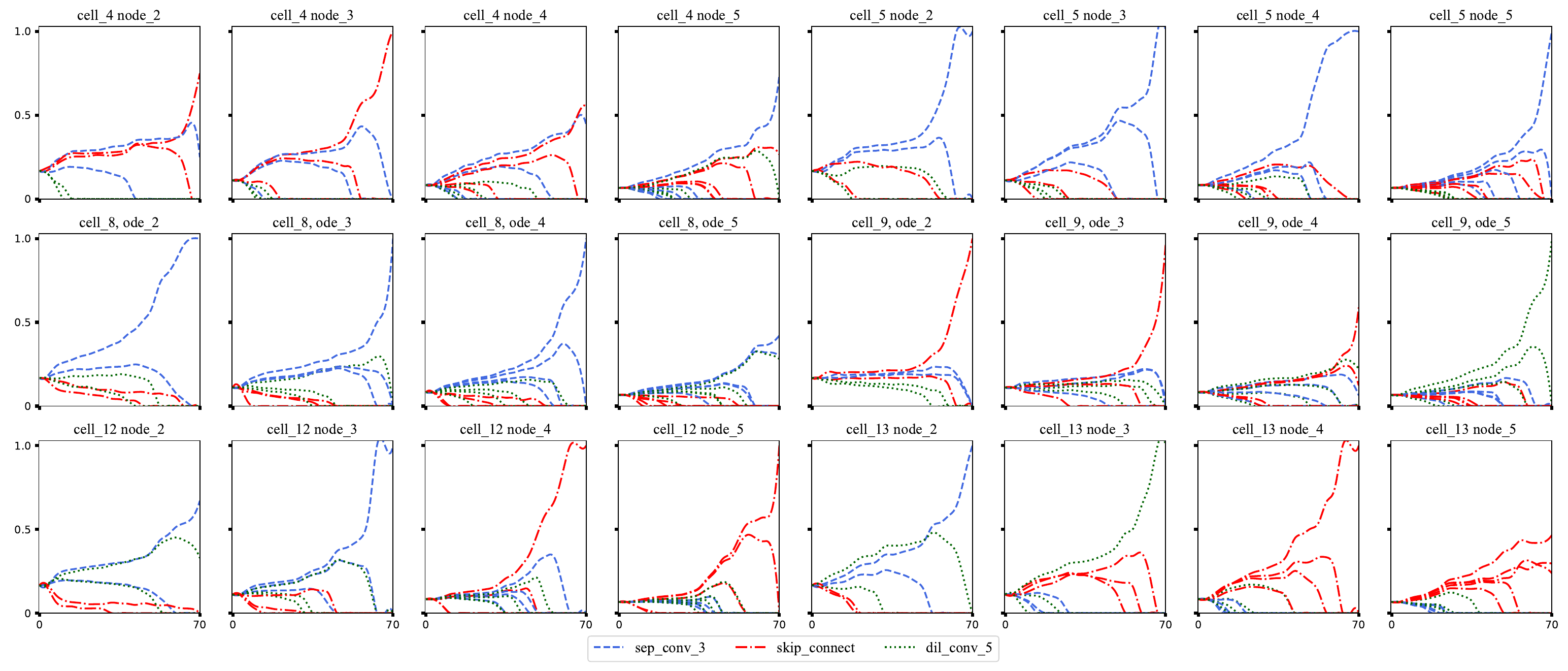}}
	\caption{The evolution of attention weights of different operators across the search procedure (for brevity, only cells 4, 8 and 12 are presented)}
	\label{fig:architecture_weights}
\end{figure*}
Like GOLD-NAS, a notable benefit of the proposed SNPS and DNAD lies in that a set of Pareto-optimal architectures can be derived within a search procedure. Specifically, SNPS takes 0.5 GPU-days to complete the search process using an RTX 3090 card, and DNAD takes a longer time of 0.6 GPU-days due to the extra computational burden on collecting feature maps of the teacher network. The efficiency for implementing DNAD is comparable with DARTS-based methods. The results are summarized in Table \ref{tab:cifar10_results}. In a search procedure, four neural architectures are picked and retrained from scratch for three times to evaluate their final performances. It takes about 0.75 GPU-days to evaluate an architecture on CIFAR-10 using a RTX 3090 card. 

The evaluation results are summarized in Table \ref{tab:cifar10_results}. Specifically, we prefer architectures with lower test error rates or fewer parameters. SNPS-Net-R$_{1-4}$ and DNAD-Net-R$_{1-4}$ are architectures searched by SNPS and DNAD, respectively. It can be seen that both the SNPS and DNAD achieve competitive NAS results on CIFAR-10. In particular, the SNPS-Net-R$_{4}$, with 3.5M parameters, achieves an error rate of $2.53\%$, which is a competitive result to other NAS baselines. The SNPS-Net-R$_{1}$, with only 2.2M parameters, achieves an error rate of $2.73\%$, which demonstrates that the SNPS is able to derive architectures with parameter efficiency. Furthermore, with the guidance of teacher knowledge, the DNAD derives a better Pareto front than SNPS. In particular, the DNAD-Net-R$_{4}$, with 3.3M parameters, achieves an error rate of $2.47\%$. And the DNAD-Net-R$_{1}$, with only 2.1M parameters, achieves an error rate of $2.69\%$. Partial architectures searched by DNAD are visualized in Fig. \ref{fig:NADD_nets}. 


It is worth pointing out that the flexibility of the neural architectures of our algorithms assists to derive the architecture set which strikes appropriate tradeoffs between the model performance and the computational complexity. In conventional DARTS-based approaches, deleting an operator in a neural cell means deleting this operator in all cells of the same type. In the developed SNPS and DNAD, deleting an operator in a neural cell causes little interference in the performance of the dynamic super-network. In this way, an intermediate super-network can be viewed as a favorable successor of the previous super-network. 
\subsection{Search procedure and Grad-CAM visualization}
\begin{figure}[htbp]
	\centering
	\includegraphics[width=0.80\linewidth]{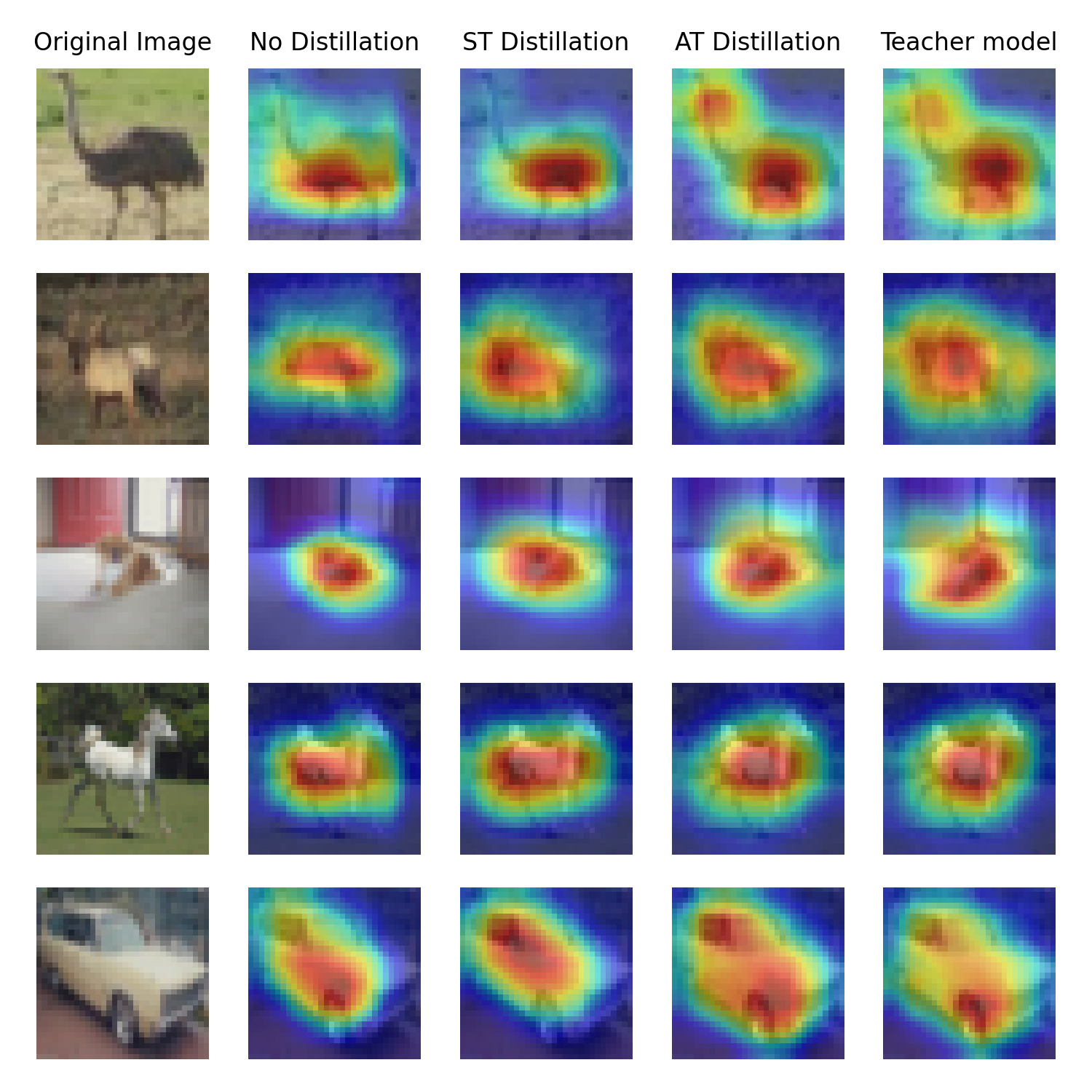}
	\caption{Grad-CAM visualization with different KD settings}
	\label{fig:gradcamvisualization}
\end{figure}

\paragraph{Search procedure} The overall fluctuations of multiplicative sparsity coefficient $\gamma$ and additive sparsity coefficient $\mu$ during the search procedure are visualized in Fig. \ref{fig:search_procedure}. In the early stage, $\gamma$ is larger than $\gamma\mu$, which implies that the super-network pays more attention to decrease the performance-attentive sparsity loss until the super-network is not sparse enough as expectation. The sparsity entropy $\mathcal{L}_{S}(\alpha)$ decreases constantly during the search procedure. The weight evolution of some operators are also visualized in Fig. \ref{fig:architecture_weights}. We can see that the weights of some unimportant operators are suppressed to zero gradually during the search procedure. In this way, the super-network is progressively shrunk from a dense and bloated structure to a sparse and lightweight one. 

\paragraph{Grad-CAM visualization} In Fig. \ref{fig:gradcamvisualization}, the gradient weighted class activation mapping (Grad-CAM) \cite{selvaraju2020grad} is visualized for several images in CIFAR-10 with different KD settings. As it is shown, the activation maps generated by the AT-distilled super-network have similar spatial patterns and areas of high activation as those of the teacher model. It indicates that AT distillation is effective in transferring the knowledge, especially in terms of the feature distribution. In contrast, the activation maps of the ST-distilled model are more similar to those of the model which does not use KD. It indicates that the ST distillation is not able to achieve stable knowledge transfer from the teacher model to the student super-network. 

\subsection{Generalization to SVHN dataset}
\begin{figure}[htbp]
	\centering
	\subfloat[$\beta$-DARTS]{
		\includegraphics[width=0.30\linewidth]{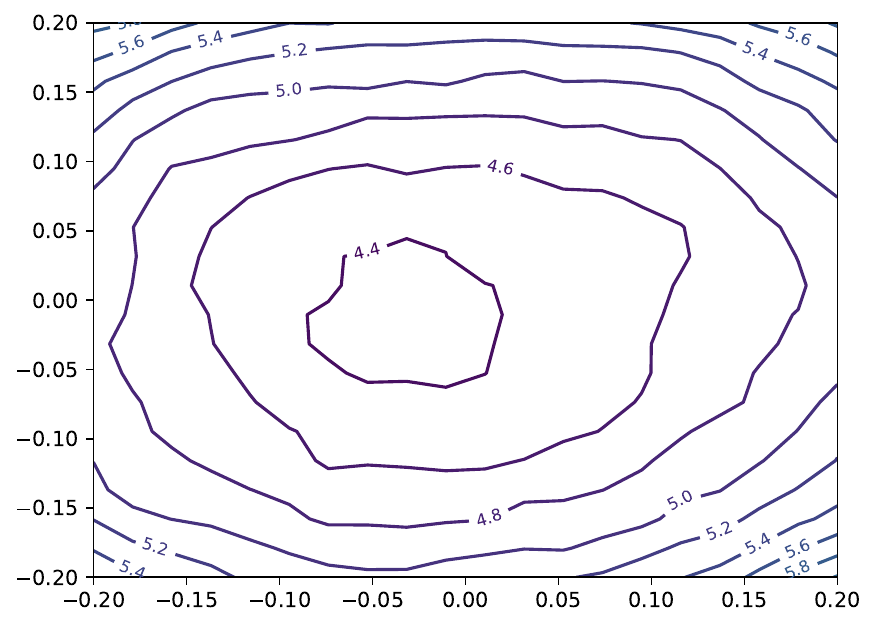}}
	\subfloat[CDARTS]{
		\includegraphics[width=0.30\linewidth]{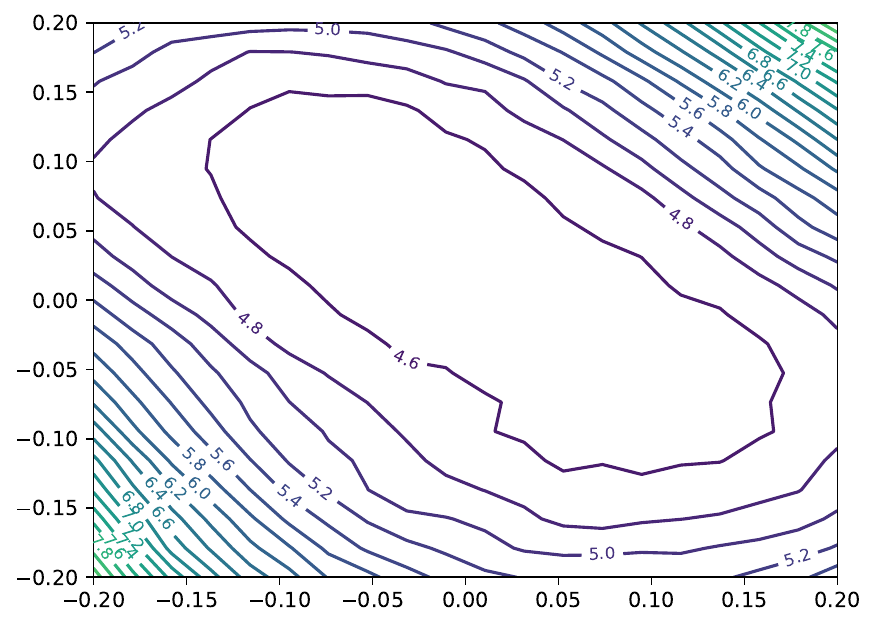}} 
	\subfloat[DOTS]{
		\includegraphics[width=0.30\linewidth]{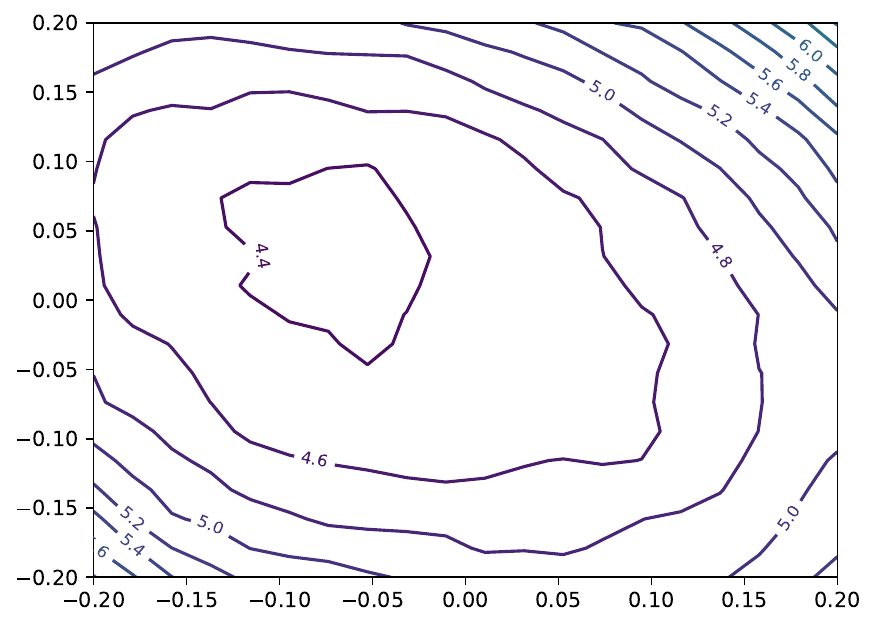}} \\
	\subfloat[PC-DARTS]{
		\includegraphics[width=0.30\linewidth]{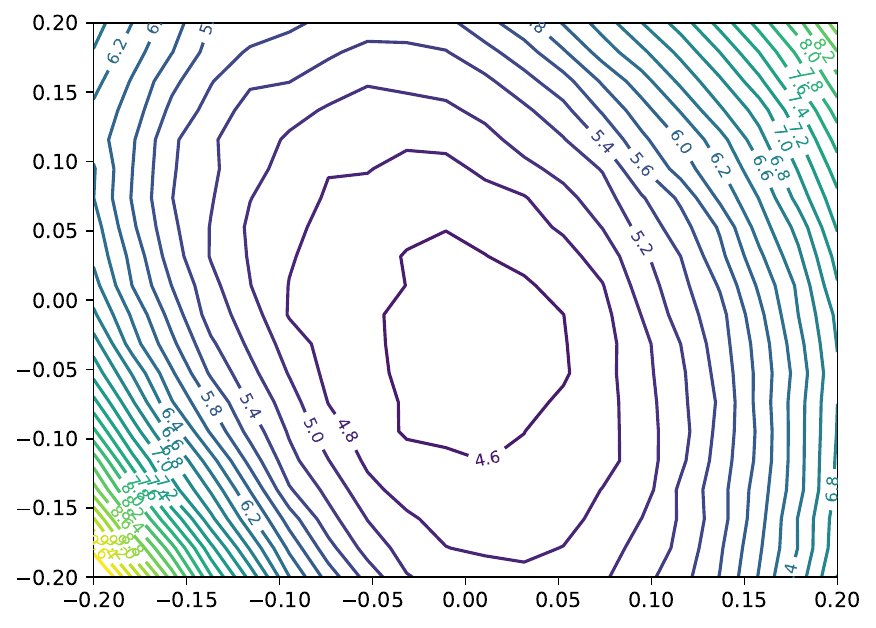}}	\subfloat[DNAD-Net-R$_{1}$]{
		\includegraphics[width=0.30\linewidth]{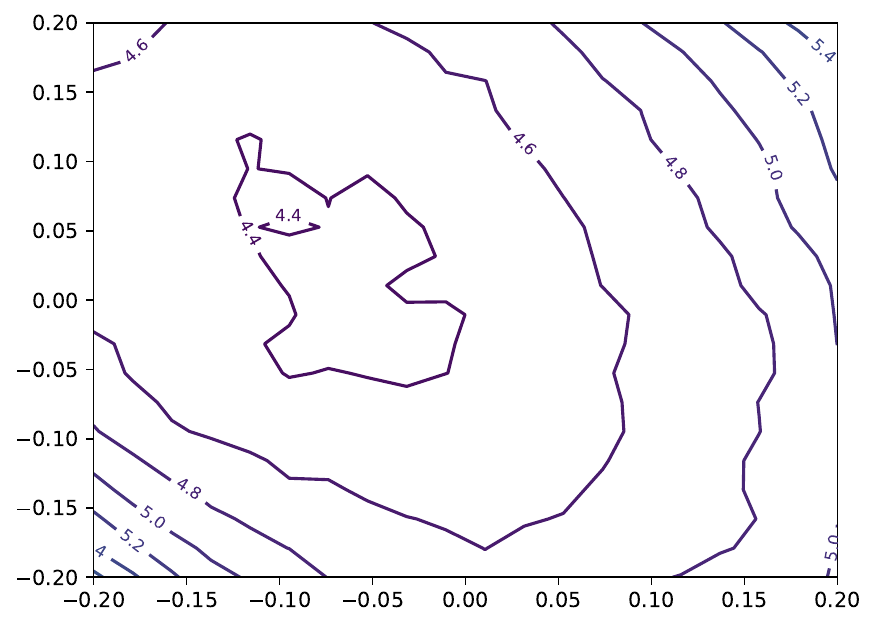}}
	\subfloat[DNAD-Net-R$_{4}$]{
		\includegraphics[width=0.30\linewidth]{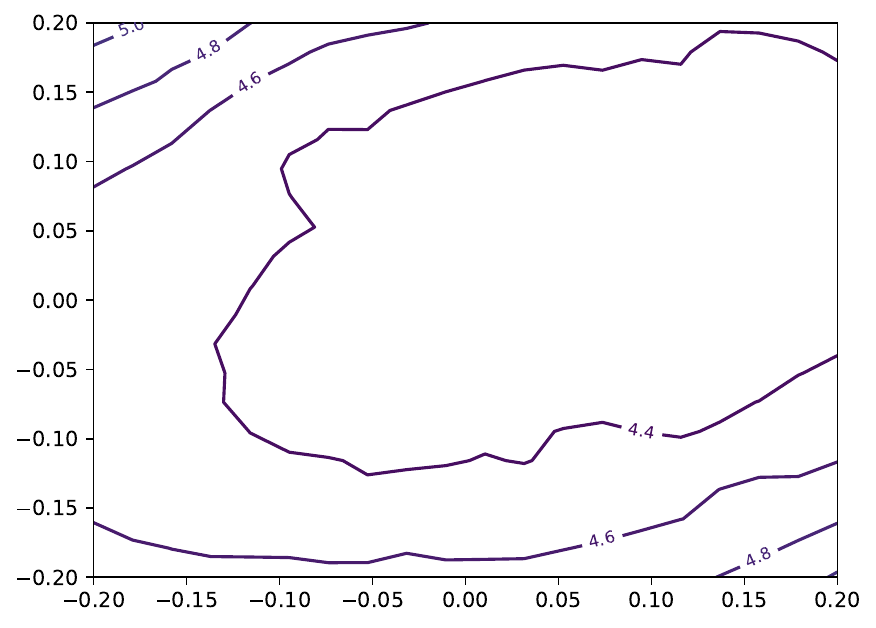}}
	\caption{Contour plots of test error rate in 2D random subspaces on the SVHN dataset. Specifically, let $\bm{w}^{*} \in \mathbb{R}^{d}$ be the converged weight of a network, and $\bm{d}_{1}, \bm{d}_{2} \in \mathbb{R}^{d}$ are random directions with filter normalization. We test the error rate of network with modified weight $\bm{w}^{*} = l_{1}\bm{d}_{1} + l_{2}\bm{w}$, where $l_{1}$ and $l_{2}$ are the coordinates of subspace.}
	\label{fig:losssurface}
\end{figure}

\begin{table}[htbp]
	\centering
	\caption{Evaluation results of generalization experiments on SVHN}
	\label{tab:SVHN_results}
	\scalebox{1.0}{
	\begin{tabular}{l|c|c}
		\toprule
	 Architecture & Params {(M)} & Test Err. (\%) \\
	 \midrule
	 $\beta$-DARTS \cite{DBLP:conf/cvpr/YeL00FO22} & 3.6 & 4.41 $\pm$ 0.12 \\
	 CDARTS \cite{9720178} & 3.9 & 4.41 $\pm$ 0.13 \\
	 DARTS(2nd) \cite{liu2018darts} & 3.3 & 4.35 $\pm$ 0.06 \\
	 DOTS \cite{gu2021dots} & 3.5 & 4.30 $\pm$ 0.09 \\
	 PC-DARTS \cite{xu2019pc} & 3.6 & 4.57 $\pm$ 0.14 \\
	 \midrule
	 DNAD-Net-R$_{1}$ & 2.1 & 4.43 $\pm$ 0.03 \\
	 DNAD-Net-R$_{4}$ & 3.3 & 4.31 $\pm$ 0.05 \\
	 \bottomrule
	\end{tabular}
}
\end{table}
In this part, we evaluate the classification accuracy of partial architectures searched on CIFAR-10 on the SVHN dataset to test their generalization performance in other domains. Specifically, we randomly choose 10,000 images from the training set to form a new set, which makes a network more susceptible to over-fitting. The networks are trained using an SGD optimizer with an initial learning rate that anneals to zero over 200 epochs. Other data enhancement techniques include cutout of size $16\times16$ and drop-path with a rate of 0.2 for DARTS-like architectures and 0.3 for DNAD architectures. All architectures are evaluated three times with different random seeds. The performance of an architecture is evaluated by the classification accuracy on the validation set of SVHN, which contains 26,032 images and 10 classes. The evaluation results are summarized in Table \ref{tab:SVHN_results}. As can be seen, DNAD-Net-R$_{1}$, which has only 2.1M parameters, achieves better results than the more computationally expensive PC-DARTS.

To evaluate the generalization performance across different architectures, we employ a loss landscape analysis method following \cite{DBLP:conf/nips/Li0TSG18}. For each trained network, 2D subspaces are constructed by sampling random perturbation vectors $\bm{d}_1$ and $\bm{d}_2$ that are normalized to match filter-wise magnitudes of converged weights $\bm{w}^*$. Test errors are then computed and visualized in Fig. \ref{fig:losssurface} by employing perturbed weights $\bm{w} = \bm{w}^* + l_1\bm{d}_1 + l_2\bm{d}_2$, where $\bm{w}^*$ is the converged weights of network training. Specifically, the geometric flatness of loss landscapes directly correlates with generalization robustness. As shown in Fig. \ref{fig:losssurface}, DNAD-Net-R$_4$ exhibits broad, flat minima where parameter perturbations (e.g., $l_1,l_2$ variations) cause minimal performance degradation (less than 0.5\% test error fluctuation). In contrast, sharp basins observed in $\beta$-DARTS and PC-DARTS reveal the sensitivity to weight shifts, with error spikes exceeding 2\% under equivalent perturbations. Architectural differences manifest distinctly in these subspaces. While topology-constrained methods like CDARTS show fragmented contours with local minima, DNAD's hierarchical feature alignment produces smoother transitions among modes, which explains its superior SVHN performance (4.31\% vs. CDARTS’s 4.41\%) with 18\% fewer parameters.

\section{Experimental Results on ImageNet}
\begin{figure}[htbp]	
	\centering{\includegraphics[width=0.50\textwidth]{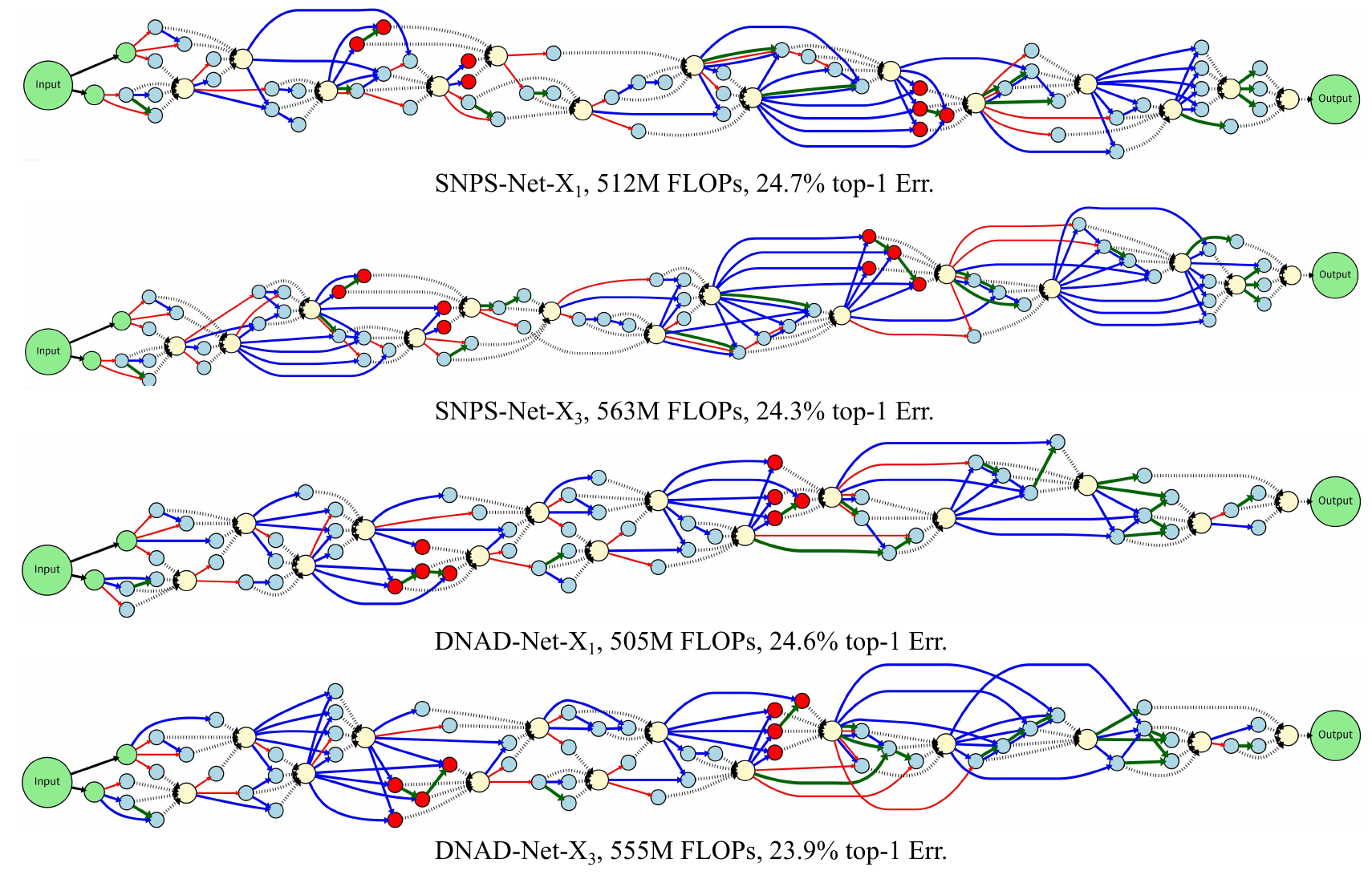}}
	\caption{Neural architectures searched by DNAD on ImageNet. In this figure, neural architectures are named in a sescending sequence of model size.}
	\label{fig:NADD_Nets_X}
\end{figure}
In this section, we evaluate the performance of SNPS and DNAD on the ImageNet classification task.
\subsection{Hyper-parameter settings}
\subsubsection{Super-network}
Most of the super-network settings in ImageNet experiments are the same as those in CIFAR-10. The exceptions are listed as follows. The batch size is set to be 256, and no data-augment techniques are employed since the ImageNet is large enough to avoid over-fitting. 
\subsubsection{SNPS}
The initial multiplicative sparsity coefficient $\gamma_{0}$ is set to be $1\times 10 ^{-6}$. The upper bound $\gamma_{\rm max}$ is set to be $1\times 10^{-4}$. The moving average coefficient $\rho$ is $0.99$. The expected prune number $n_{\rm prune}$ is set to be $2\times10^{-4}$. Other hyper-parameter settings are the same as those in CIFAR-10. 
\subsubsection{DNAD}
It is inefficient to train a teacher network from scratch on ImageNet, which contains 1.3M training and 50K testing images. Thus, a pre-trained EfficientNet-B0 \cite{tan2019efficientnet} is selected as our teacher network. In particular, the pre-trained model achieves an error rate of 23.7$\%$ on ImageNet. Similar to previous section, we divide the teacher network into $M=3$ blocks and collect the corresponding feature maps to calculate the AT knowledge loss. 
\subsubsection{Discrete architectures} For each search procedure, four neural architectures are picked and re-trained on ImageNet. In particular, each architecture is re-trained for 250 epochs with a batch size of 256. The channel number is set to be 48. The network parameters is optimized by an SGD optimizer with an initial learning rate of 0.25 and a momentum of 0.9. The learning rate is annealed to zero following a linear scheduler. The weight decay is set to be $3\times10^{-5}$. The label smoothing and an auxiliary loss of 0.4 probability are employed as additional enhancements. The learning rate warm-up is applied to the first 5 epochs. 
\subsection{Evaluation results}
Both SNPS and DNAD are implemented on a single A6000 card, and it takes 1.3 or 1.5 GPU-days to complete the architecture search. The searched architectures are trained by three A6000 cards in parallel for acceleration. The evaluation results of SNPS and DNAD are compared with recent state-of-the-art NAS approaches in Table \ref{tab:imagenet_results}. The searched architectures by DNAD are visualized in Fig. \ref{fig:NADD_Nets_X}. We prefer architectures with lower test error rates and fewer parameters and FLOPs. As we can see, both SNPS and DNAD show competitive performance among recent state-of-the-art NAS approaches. In particular, the SNPS-Net-X$_{4}$, the architecture searched without any knowledge guidance, achieves an error rate of $23.9\%$ with 597M FLOPs, which is comparable with GOLD-NAS-Z (23.9$\%$ error with 585M FLOPs) and is higher than other NAS baselines. Furthermore, the DNAD-Net-X$_{4}$, which is searched under the guidance of a teacher network, achieves an error rate of $23.7\%$ with 598M FLOPs, which is higher than the original SNPS-Net-X$_{4}$. The other smaller architectures searched by SNPS and DNAD are also competitive. For example, the SNPS-Net-X$_{1}$, achieves an error rate of $24.6\%$ with only 5.1M parameters and 505M FLOPs. A comparable architecture is the one searched by MiLeNAS, which achieves an error rate of $24.7\%$. However, the MiLeNAS architecture, which owns 5.3M parameters and 585M FLOPS, is much larger than SNPS-Net-X$_{1}$. Furthermore, the DNAD-Net-X$_{1}$, achieves an error rate of $24.6\%$ with 5.1M parameters and 505M FLOPS, which is slightly better than SNPS-Net-X$_{1}$ and much higher than other architectures with comparable computational burdens. Overall, both the developed SNPS and DNAD are competitive to the listed NAS baselines. 
\begin{table}[htbp]
	\centering
	\caption{Comparison with recent state-of-the-art NAS approaches on ImageNet}
	\label{tab:imagenet_results}
	\scalebox{0.90}{
	\begin{tabular}{l|c|c|c|c}
		\toprule
		\multirow{2}{*}{{Architecture}} & {Params} & {FLOPs} & {Top-1 Err.} & {Search cost} \\
		{} & {(M)} & {(M)} &($\bm{\%}$) & {(GPU-days)} \\
		\midrule
		NASNet-A \cite{zoph2018learning} & 5.3 & 564 &  26.0 & 1800 \\
		DARTS($2^{\rm nd}$) \cite{liu2018darts} & 4.7 & 574 & 26.7 & 4.0 \\
		SNAS \cite{xie2018snas} & 4.3 & 522 & 27.3 & 1.5 \\
		P-DARTS \cite{chen2019progressive} & 4.9 & 557 & 24.4 & 0.3 \\
		PC-DARTS \cite{xu2019pc} & 5.3 & 597 & 24.2 & 0.1 \\ 
		GDAS(FRC) \cite{dong2019searching} & 4.4 & 497 & 27.5 & 0.2 \\ 		
		DA$^{2}$S \cite{tian2021discretization} & 5.0 & 565 & 24.4 & 0.3 \\
		CDARTS(MS) \cite{9720178} & 5.4 & 571 & 24.1 & 1.7 \\
		CDARTS \cite{9720178} & 6.1 & 701 & 23.7 & 1.7 \\
		MiLeNAS \cite{he2020milenas} & 5.3 & 584 & 24.7 & 0.3 \\ 
		RelativeNAS \cite{9488309} & 5.03 & 563 & 24.9 & 0.4 \\
		PWSNAS \cite{9739130} & - & 645 & 24.1 & - \\
		PWSNAS \cite{9739130} & - & 570 & 24.4 & - \\
		DOTS(ImageNet) \cite{gu2021dots} & 5.3 & 596 & 24.0 & 1.3 \\
		LFR-DARTS \cite{DBLP:journals/soco/HaoZ22} & 4.9 & - & 25.5 & 0.45 \\ 
		DistillDARTS-SD \cite{DBLP:conf/icmcs/LiaoZWYLRYF22} & 5.1 & 590 & 25.6 & 2.3 \\
		DistillDARTS-SD \cite{DBLP:conf/icmcs/LiaoZWYLRYF22} & 5.5 & 636 & 24.2 & 0.5 \\ 
		$\beta$-DARTS \cite{DBLP:conf/cvpr/YeL00FO22} & 5.5 & 609 & 23.9 & 0.4 \\
		EoiNAS \cite{9432795} & 5.0 & 570 & 25.6 & 0.6 \\
		ReCNAS(w/o SE) \cite{9836970} & 6.2 & - & 23.9 & - \\
		PDARTS-AER \cite{jing2023architecture} & 5.1 & 578 & 24.0 & 2.0 \\
		PADARTS \cite{xue2023improved} & 5.2 & - & 24.7 & 0.4 \\
		$\ell$-DARTS \cite{hu2024} &  5.8 & - & 24.9 & 0.1 \\
		SWD-NAS \cite{xue2024self} & 6.3 & - & 24.5 & 0.1 \\
		STO-DARTS-V2 \cite{cai2024sto} & 3.8 & - & 24.9 & 0.4 \\
		SaDENAS \cite{han2024sadenas} & 5.6 & - & 24.9 & - \\
		EG-NAS \cite{cai2024eg} & 5.3 & - & 24.9 & 0.1 \\
		OLES \cite{jiang2024operation} & 4.7 & - & 24.5 & - \\
		DARTS-PT-CORE \cite{xie2024darts} & 5.0 & - & 25.0 & 0.8 \\
		
		\midrule
		Gold-NAS-X \cite{bi2020gold} & 6.4 & 585 & 24.3 & 2.5 \\
		Gold-NAS-Y \cite{bi2020gold} & 6.4 & 578 & 24.3 & 2.1 \\
		Gold-NAS-Z \cite{bi2020gold} & 6.3 & 585 & 24.0 & 1.7 \\
		\midrule
		{SNPS-Net-X$_{1}$} & 5.1 & 512 & 24.7 & \multirow{4}{*}{1.3} \\
		{SNPS-Net-X$_{2}$} & 5.6 & 530 & 24.4 & {} \\
		{SNPS-Net-X$_{3}$} & 5.7 & 563 & 24.1 & {} \\
		{SNPS-Net-X$_{4}$} & 6.0 & 597 & 23.9 & {} \\
		\midrule
		{DNAD-Net-X$_{1}$} & 5.1 & 505 & 24.6 & \multirow{4}{*}{1.5} \\
		{DNAD-Net-X$_{2}$} & 5.3 & 523 & 24.3 & {} \\
		{DNAD-Net-X$_{3}$} & 5.6 & 555 & 23.9 & {} \\
		{DNAD-Net-X$_{4}$} & 6.0 & 598 & 23.7 & {} \\
		\bottomrule
	\end{tabular}}
\end{table}

\section{Ablation Experiments and Discussion}
\subsection{What distillation method matters?}
In this part, the effect of different distillation methods is investigated by using AT knowledge loss \eqref{eq:knowledge_loss}, ST knowledge loss \eqref{eq:st_loss} and SA knowledge loss (ST + AT) \eqref{eq:sa_loss}. In addition, we try to test whether the regularization brought by KD is more powerful than data-augment techniques. Specifically, three settings, which are none, auto-augment and cutout, are tested. Furthermore, we also combine these techniques with KD to test whether KD + data-augment is better than single KD. 

\begin{table}[htbp]
	\centering
	\caption{Ablation experimental results with different distillation methods}
	\label{tab:distillation_methods}
	\scalebox{0.70}{
		\begin{tabular}{c|c|c|c|c|c|c}
			\toprule
			{Distillation} & {Data-augment} & {Architecture} & {Params(M)} & 
			{FLOPs(M)} & {Test Err. (\%)} & {Comp.}  \\
			\midrule
			\multirow{3}{*}{None} & \multirow{3}{*}{None} & SNPS-NN-A & 0.322 & 62.27 & 4.93 $\pm$ 0.08 & - \\ 
			{} & {} & SNPS-NN-B & 0.518 & 92.45 & 4.28 $\pm$ 0.10 & -\\
			{} & {} & SNPS-NN-C & 0.622 & 109.1 & 4.10 $\pm$ 0.04 & -\\
			\midrule
			\multirow{3}{*}{None} & \multirow{3}{*}{Auto-augment} & SNPS-NA-A & 0.360 & 66.97 & 4.69 $\pm$ 0.09 & 0.24$\downarrow$\\
			{} & {} & SNPS-NA-B & 0.506 & 88.47 & 4.20 $\pm$ 0.12 & 0.08$\downarrow$ \\
			{} & {} & SNPS-NA-C & 0.600 & 108.7 & 4.16 $\pm$ 0.11 & 0.06$\uparrow$ \\
			\midrule
			\multirow{3}{*}{None} & \multirow{3}{*}{Cutout} & SNPS-NC-A & 0.336 & 64.15 & 4.74 $\pm$ 0.05 & 0.19$\downarrow$ \\
			{} & {} & SNPS-NC-B & 0.472 & 89.70 & 4.28 $\pm$ 0.21 & 0.00 \\
			{} & {} & SNPS-NC-C & 0.574 & 107.2 & 4.06 $\pm$ 0.07 & 0.04$\downarrow$ \\
			\midrule
			\multirow{3}{*}{ST} & \multirow{3}{*}{None} & DNAD-SN-A & 0.394 & 68.01 & 4.77 $\pm$ 0.15 & 0.16$\downarrow$ \\
			{} & {} & DNAD-SN-B & 0.549 & 92.06 & 4.28 $\pm$ 0.18 & 0.00 \\
			{} & {} & DNAD-SN-C & 0.658 & 109.4 & 4.10 $\pm$ 0.02 & 0.00 \\ 
			\midrule
			\multirow{3}{*}{ST} & \multirow{3}{*}{Auto-augment} & DNAD-SA-A & 0.436 & 68.79 & 4.90 $\pm$ 0.08 & 0.03$\downarrow$\\
			{} & {} & DNAD-SA-B & 0.603 & 91.32 & 4.42 $\pm$ 0.25 & 0.14$\uparrow$\\
			{} & {} & DNAD-SA-C & 0.687 & 108.1 & 4.01 $\pm$ 0.10 & 0.09$\downarrow$\\
			\midrule
			\multirow{3}{*}{ST} & \multirow{3}{*}{Cutout} & DNAD-SC-A & 0.406 & 69.71 & 4.63 $\pm$ 0.02 & 0.30$\downarrow$ \\
			{} & {} & DNAD-SC-B & 0.564 & 91.38 & 4.30 $\pm$ 0.04 & 0.02$\uparrow$\\
			{} & {} & DNAD-SC-C & 0.653 & 110.0 & 4.21 $\pm$ 0.13 & 0.11$\uparrow$ \\
			\midrule
			\multirow{3}{*}{AT} & \multirow{3}{*}{None} & DNAD-AN-A & 0.458 & 68.33 & 4.83 $\pm$ 0.10 & 0.10$\downarrow$ \\
			{} & {} & DNAD-AN-B & 0.749 & 91.34 & 4.23 $\pm$ 0.08 & 0.05$\downarrow$\\
			{} & {} & DNAD-AN-C & 0.847 & 107.0 & 3.95 $\pm$ 0.17 & 0.15$\downarrow$\\
			\midrule
			\multirow{3}{*}{AT} & \multirow{3}{*}{Auto-augment} & DNAD-AA-A & 0.422 & 67.73 & 4.68 $\pm$ 0.09 & 0.25$\downarrow$ \\
			{} & {} & DNAD-AA-B & 0.668 & 88.74 & 4.17 $\pm$ 0.09 & 0.11$\downarrow$ \\
			{} & {} & DNAD-AA-C & 0.862 & 109.8 & 4.05 $\pm$ 0.06 & 0.05$\downarrow$\\
			\midrule
			\multirow{3}{*}{AT} & \multirow{3}{*}{Cutout} & DNAD-AC-A & 0.406 & 64.88 & 4.64 $\pm$ 0.17 & 0.29$\downarrow$ \\
			{} & {} & DNAD-AC-B & 0.564 & 91.38 & 4.15 $\pm$ 0.06 & 0.13$\downarrow$ \\
			{} & {} & DNAD-AC-C & 0.783 & 110.0 & 3.92 $\pm$ 0.04 & 0.18$\downarrow$ \\        
			\midrule
			\multirow{3}{*}{ST + AT} & \multirow{3}{*}{None} & DNAD-SAN-A & 0.406 & 69.79 & 4.63 $\pm$ 0.03 & 0.30$\downarrow$ \\
			{} & {} & DNAD-SAN-B & 0.591 & 90.39 & 4.12 $\pm$ 0.12 & 0.16$\downarrow$ \\
			{} & {} & DNAD-SAN-C & 0.682 & 106.2 & 4.09 $\pm$ 0.07 & 0.01$\downarrow$ \\
			\midrule
			\multirow{3}{*}{ST + AT} & \multirow{3}{*}{Auto-augment} & DNAD-SAA-A & 0.422 & 67.55 & 4.68 $\pm$ 0.11 & 0.25$\downarrow$\\
			{} & {} & DNAD-SAA-B & 0.590 & 89.76 & 4.27 $\pm$ 0.02 & 0.01$\downarrow$\\
			{} & {} & DNAD-SAA-C & 0.705 & 108.2 & 3.92 $\pm$ 0.06 & 0.18$\downarrow$ \\
			\midrule
			\multirow{3}{*}{ST + AT} & \multirow{3}{*}{Cutout} & DNAD-SAC-A & 0.431 & 70.95 & 4.37 $\pm$ 0.09 & 0.56$\downarrow$ \\
			{} & {} & DNAD-SAC-B & 0.583 & 89.73 & 4.11 $\pm$ 0.02 & 0.17$\downarrow$ \\
			{} & {} & DNAD-SAC-C & 0.712 & 110.4 & 3.95 $\pm$ 0.07 & 0.15$\downarrow$ \\           
			\midrule
		\end{tabular}
	}
\end{table}

To make the ablation study more comprehensive, we search smaller neural architectures to allow more evaluations. In particular, the super-network consists of 11 neural cells and the initial channel is 16. For each search procedure, 3 neural architectures are picked and re-trained for 200 epochs. The initial channel of these discrete architectures is also 16. The probability of path dropout is set to be 0.2. Other hyper-parameter settings are the same as CIFAR-10 evaluations. To obtain a fair comparison, we restrict the FLOPs of searched architecture by three levels, $65.00$M, $90.00$M and $110.0$M, respectively. For each level, the architecture searched by SNPS without any regularization term is regarded as the baseline. The evaluation results are summarized in Table \ref{tab:distillation_methods}. In particular, `Comp.' means the comparison with the baseline.

From Table \ref{tab:distillation_methods}, we can see that both the AT and ST + AT improve the performance of SNPS significantly. In particular, without data-augment techniques, the average error rates achieved by DNAD are significantly lower than the baselines (see SNPS-NN vs. DNAD-AN: 4.93$\%$ vs. 4.83$\%$, 4.28$\%$ vs. 4.23$\%$, and 4.10$\%$ vs. 3.95$\%$). The data-augment techniques, such as auto-augment and cutout, produce slight improvements on the performance of SNPS, but the improvements are not obvious compared to KD techniques such as AT and ST + AT. With data-augment techniques, the DNAD performs better than the naive version of DNAD. For example, with the cutout, the DNAD-AC achieves better performance than DNAD-AN in all three levels. It implies that the combination of data-augment techniques and KD methods bring more powerful regularization impacts on NAS. 

However, it is surprising that the ST knowledge does not improve the performance of SNPS, and it achieves even worse performance than the original SNPS occasionally (see DNAD-SC). Similar phenomenons are observed in ST + AT knowledge loss. For example, with ST and AT, the DNAD achieves similar performance as DNAD with AT. It implies that KD does not always improve the NAS results. One potential reason is that the ST only utilizes the final predictive probabilities to guide the super-network optimization, while the AT utilizes informative attention maps to factorizes the learning process into step-by-step imitation. 

\subsection{DNAD in unstable operator spaces}  
In this part, we investigate the performance of DNAD in two unstable operator spaces. The GOLD-NAS has tried searching neural architectures in an operator space which includes the max-pooling. However, the architectures searched by GOLD-NAS in such an operator space are governed by the max-pooling and achieve unsatisfactory performances. In this section, two additional operator spaces, which include the max-pooling and average-pooling additionally, are implemented. The AT knowledge loss and cutout are used. Other hyper-parameters remain the same as CIFAR-10 evaluations. 
\begin{table}[htbp]
	\centering
	\caption{Evaluation results of SNPS and DNAD in unstable operator spaces }
	\label{tab:DNAD_pool}
	\scalebox{0.8}
	{
	\begin{tabular}{l|c|c|c|c}
		\toprule
		\multirow{2}{*}{{Architecture}} & {Params} & {FLOPs} &{Top-1 Err.} & \multirow{2}{*}{{Comp.}} \\
		{} & {(M)} & {(M)} &($\bm{\%}$) \\
		\midrule
		SNPS-Net-MaxP-A & 0.424 & 72.03 & 4.72 $\pm$ 0.14 & -\\
		SNPS-Net-MaxP-B & 0.504 & 90.54 & 4.65 $\pm$ 0.09 & -\\
		SNPS-Net-MaxP-C & 0.608 & 110.3 & 4.35 $\pm$ 0.11 & -\\
		\midrule
		DNAD-Net-MaxP-A & 0.473 & 69.36 & 4.66 $\pm$ 0.12 & 0.06$\downarrow$\\
		DNAD-Net-MaxP-B & 0.652 & 86.90 & 4.48 $\pm$ 0.07 & 0.17$\downarrow$\\
		DNAD-Net-MaxP-C & 0.804 & 107.9 & 4.09 $\pm$ 0.08 & 0.26$\downarrow$\\
		\midrule
		SNPS-Net-AvgP-A & 0.338 & 66.02 & 4.78 $\pm$ 0.11 & - \\
		SNPS-Net-AvgP-B & 0.467 & 91.92 & 4.45 $\pm$ 0.11 & - \\
		SNPS-Net-AvgP-C & 0.523 & 110.7 & 4.29 $\pm$ 0.15 & - \\
		\midrule
		DNAD-Net-AvgP-A & 0.394 & 64.94 & 4.66 $\pm$ 0.14 & 0.12$\downarrow$\\
		DNAD-Net-AvgP-B & 0.678 & 90.36 & 4.20 $\pm$ 0.05 & 0.25$\downarrow$ \\
		DNAD-Net-AvgP-C & 0.827 & 109.1 & 4.14 $\pm$ 0.10 & 0.15$\downarrow$\\
		\bottomrule		
	\end{tabular}
}
\end{table}

\begin{figure}[htbp]
	\label{fig:unstable_search_spaces}
	\centering{\includegraphics[width=0.45\textwidth]{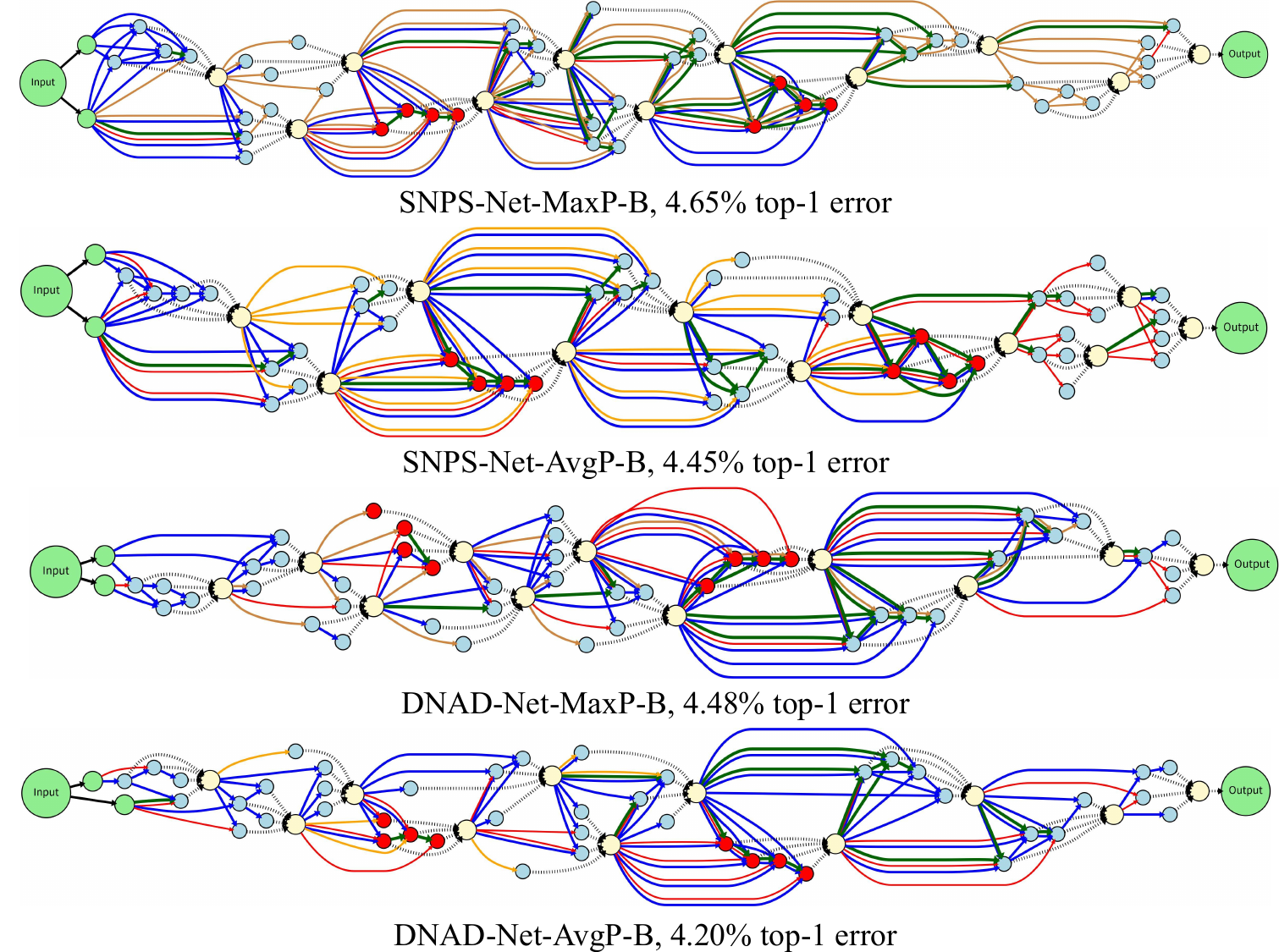}}
	\caption{Part of the architectures searched by SNPS and DNAD in unstable operator spaces. The max-pooling operator is marked by brown directed solid lines. The average-pooling operator is marked by yellow directed solid lines.}
	\label{fig:DNAD_pool}
\end{figure}

The evaluation results are summarized in Table \ref{tab:DNAD_pool}. The searched architectures are visualized in Figure 8. In Table \ref{tab:DNAD_pool}, architectures SNPS-Net-MaxP and DNAD-Net-MaxP are searched in an operator space that includes max-pooling. Architectures SNPS-Net-AvgP and DNAD-Net-AvgP are searched in an operator space that includes average-pooling. In particular, architectures SNPS-Net-MaxP and SNPS-Net-AvgP are regarded as the baselines for the operator space that includes max-pooling and average-pooling, respectively. From Table \ref{tab:DNAD_pool}, we can see that DNAD achieves better performance than SNPS under the same FLOPs. For example, DNAD-Net-MaxP-C, with 107.9M FLOPs, achieves a 0.26\% lower error rate than SNPS-Net-MaxP-C with 110.3M FLOPs. Partial architectures are visualized in Fig. \ref{fig:DNAD_pool}. An intuitive observation is that SNPS architectures preserve more max-pooling or average-pooling operators, which cause unsatisfactory performance. In contrast, with the knowledge guidance, DNAD architectures are less affected and preserve more learnable operators, such as separable and dilated convolutions. Nevertheless, the experimental results show that the negative influences caused by parameter-free max-pooling and avg-pooling cannot be completely eliminated by DNAD. This means that, even with the knowledge regularization, searching a satisfactory architecture without topology constraints on cells in a full operator space remains a severe challenge which deserves further investigation.

\subsection{Does a stronger teacher derive stronger architectures?}

\begin{table}[htbp]
	\centering
	\caption{Ablation experiments on regularization impacts of different teacher networks}
	\label{tab:teacher}
	\scalebox{0.9}{
		\begin{tabular}{l|c|c|c|c}
			\toprule
			\multirow{2}{*}{{Architecture}} & {Teacher} & {Params} & {FLOPs} & {Top-1}  \\
			{} & {Err.} ($\%$) & {(M)} & {(M)} & {Err.} ($\%$) \\
			\midrule
			DNAD-Net-T$_{1}$-A & \multirow{2}{*}{2.49} & 0.853 & 149.9 & 3.76 $\pm$ 0.04 \\
			DNAD-Net-T$_{1}$-B & {} & 1.052 & 182.1 & 3.67 $\pm$ 0.08 \\
			\midrule
			DNAD-Net-T$_{2}$-A & \multirow{2}{*}{2.83} & 0.834 & 152.6 & 3.72 $\pm$ 0.11 \\
			DNAD-Net-T$_{2}$-B & {} & 1.050 & 179.3 & 3.62 $\pm$ 0.16 \\
			\midrule
			DNAD-Net-T$_{3}$-A & \multirow{2}{*}{4.14} & 0.785 & 153.4 & 3.88 $\pm$ 0.12 \\
			DNAD-Net-T$_{3}$-B & {} & 0.934 & 182.0 & 3.72 $\pm$ 0.08 \\ 
			\midrule
			SNPS-Net-A & \multirow{2}{*}{-} & 0.737 & 153.5 & 3.95 $\pm$ 0.07 \\
			SNPS-Net-B & {} & 0.889 & 180.7 & 3.84 $\pm$ 0.13 \\
			\bottomrule		
	\end{tabular}}
\end{table}

In this part, we investigate the influence of teachers' performance. Three teachers are tested with different error rates on CIFAR-10, which are $2.49\%$, $2.83\%$, and $4.14\%$, respectively. For each search procedure, two architectures are picked and retrained for 200 epochs. A 14-cell structure is used, and the channel number is set to 20. Other settings are the same as those in the previous section. The evaluation results are summarized in Table \ref{tab:teacher}.

\begin{figure}
	\centering
	\includegraphics[width=0.7\linewidth]{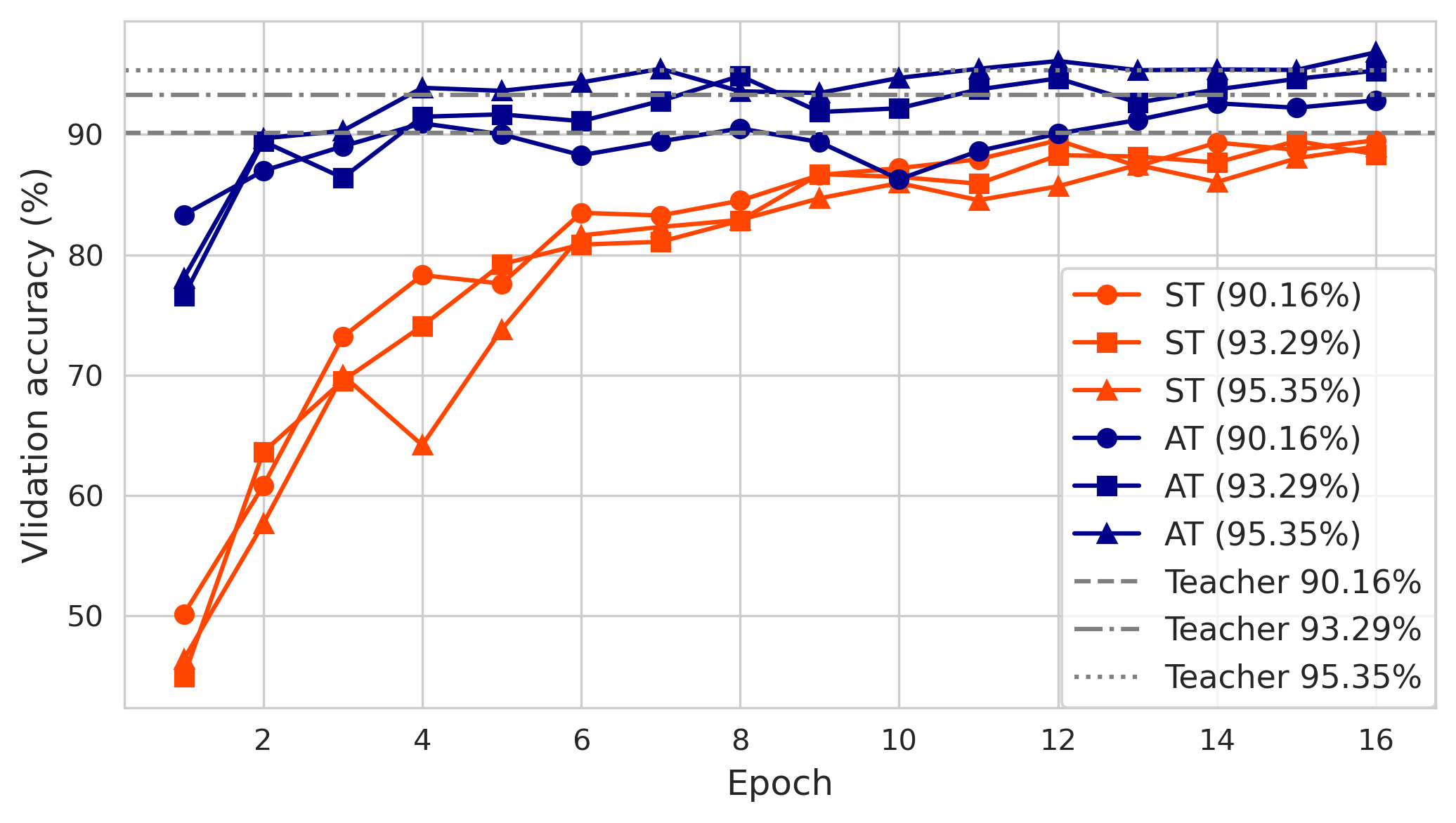}
	\caption{Validation accuracy evolution of the student super-networks on CIFAR-10 with different teachers and different KD settings}
	\label{fig:accuracycomparison}
\end{figure}

The results imply that a stronger teacher does not always derive stronger neural architectures. For example, when the teacher achieves an error rate of $2.49\%$, the derived architectures perform worse than those whose teacher achieves an error rate of $2.83\%$. This suggests that the relationship between teacher strength and architecture quality is not straightforward. Specifically, the improvement in distillation is marginal when using a strong teacher (as evidenced by DNAD-Net-T$_{1}$ and SNPS-Net-T$_{3}$, where only a 0.2\% increase is achieved despite a 1.5\% improvement in teacher performance). It implies that even an imperfect teacher can provide effective regularization during the architecture search. 

Such findings are further supported by additional experiments of training the super-networks. As shown in Fig. \ref{fig:accuracycomparison}, we visualize the accuracy evolution of super-networks on CIFAR-10 trained under different teacher supervision and KD settings during the first 16 epochs. The results indicate that the AT distillation is more effective in training the super-network, since it accelerates the convergence and enhances the performance. Notably, even when using a relatively weaker-performing teacher (90.16\%), the super-network achieves an accuracy of 93.25\%. It demonstrates that moderately strong teachers and AT distillation are sufficient to regularize the architecture search effectively. In contrast, ST distillation shows limited effectiveness in regularizing the super-network training. 
\subsection{Bayesian optimization}
\begin{figure}
	\centering
	\includegraphics[width=0.7\linewidth]{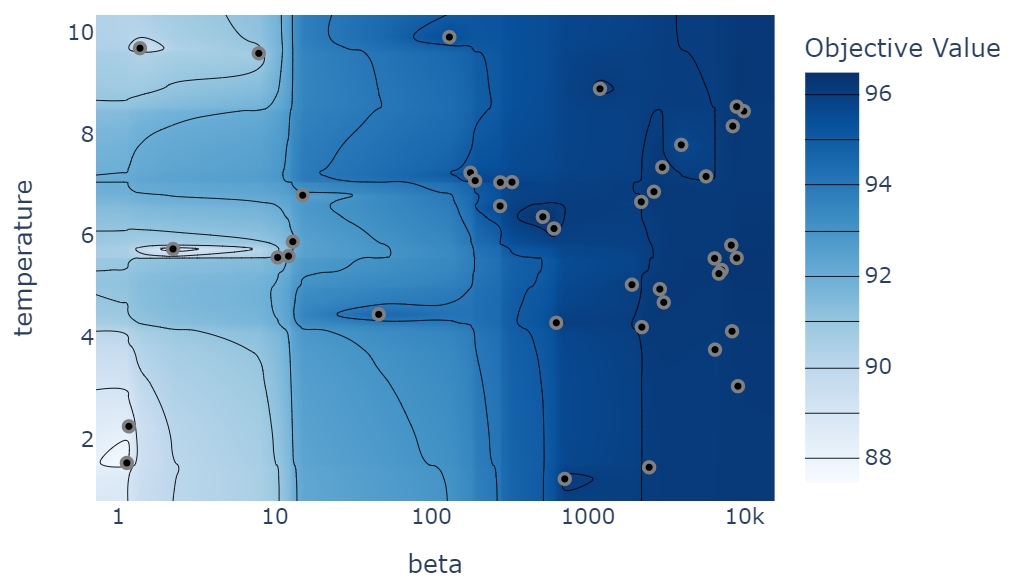}
	\caption{Performance landscape of Bayesian optimization for distillation coefficients $\beta$ and $t$ on the CIFAR-10 super-network training}
	\label{fig:optuna}
\end{figure}

In our DNAD framework implementation, certain critical hyper-parameters, such as the AT balance coefficient $\beta$ and the soft target temperature $t$, are initially adopted from their original publications. While the adaptive parameters $\gamma$ and $\mu$ are dynamically adjusted through super-network feedback mechanisms, the static nature of $\beta$ and $t$ necessitates rigorous optimization. To address this problem, a Tree-structured Parzen Estimator (TPE) based Bayesian optimization strategy via the Optuna framework is implemented for 42 experimental trials on the CIFAR-10 super-network training. 

The search space encompasses $\beta \in [1,10^4]$ (logarithmically scaled) and $t \in [1,10]$. Fig.~\ref{fig:optuna} visualizes the optimization landscape through a bivariate contour plot. As shown, the manually selected $\beta=10^3$ and $t=4$ reside in a high-performance plateau region, which provides an empirical validity despite their sub-optimal precision.
\subsection{Transfer learning study} To test the transfer ability of our searched network, we evaluate the classification performance of both the EfficientNet-B0 and DNAD-Net-X$_{4}$ on CIFAR-10 by transfer learning. As we see from Table \ref{tab:transfer_learning}, DNAD-Net-X$_{4}$ achieves a classification accuracy of 98.1$\%$ and outperforms the EfficientNet-B0 with an accuracy improvement of 0.2$\%$. 
\begin{table}[htbp]
	\centering
	\caption{Transfer learning study}
	\label{tab:transfer_learning}
	\begin{tabular}{l|c}
		\toprule
		Architecture & CIFAR-10 Acc. \\
		\midrule
		EfficientNet-B0 & 97.9$\%$ \\
		DNAD-Net-X$_{4}$ & 98.1$\%$  \\
		\bottomrule
	\end{tabular}
\end{table}

\subsection{NAS Problems that DNAD could solve}
Overall, by introducing the KD to guide the optimization of the super-network, DNAD could solve the following NAS problems. 1) It stabilizes the one-level optimization of DARTS by intermediate supervised signals and mitigates the dominance phenomenon of skip connection that has been observed widely in DARTS research. 2) It improves the performance of DARTS in a more complex search space. For example, it preserves more learnable operators in a search space where the topology is unconstrained and pooling layers are included. 3) For other NAS applications, DNAD can be utilized to accelerate the training of selected candidates, since relevant researches have demonstrated that feature-based KD boosts the fast optimization of an NN model. 

\section{Conclusions and Future Works}
In this paper, the differentiable neural architecture distillation (DNAD) algorithm is developed based on two cores, namely \emph{search by deleting} and \emph{search by imitating}. In \emph{search by deleting}, the super-network progressive shrinking (SNPS) algorithm is developed to search neural architectures in a unconstrained search space. In particular, the super-network is progressively shrunk from a dense structure to a sparse one by controlling adaptive sparsity entropy. Benefiting from the dynamic shrinking, the SNPS is able to derive a Pareto-optimal set of neural architectures within a single search procedure. In \emph{search by imitating}, the DNAD algorithm is developed by combining SNPS with KD. Since it is difficult for a student network to absorb the knowledge of a well-trained teacher network in a short time, the \emph{search by imitating} prevents the super-network from over-fitting and stabilizes the one-level optimization of DARTS. Experiments on CIFAR-10 and ImageNet demonstrate that both SNPS and DNAD are able to strike superior trade-offs between model performance computational complexity. 

Currently, the effectiveness of DNAD has been validated in AT-based and ST-based KD methods. However, its potential in exploring topology-unconstrained architectures can be further expanded by integrating it with more advanced and informative KD techniques. For instance, the relational distillation can be incorporated to enhance the knowledge transfer by leveraging structural information, such as the distances and angles between different samples \cite{park2019relational, yang2022cross, gou2022multilevel}. Additionally, pixel-wise distillation and self-supervised learning methods could be integrated to evaluate DNAD's effectiveness in alternative settings \cite{huang2023pixel}. Moreover, combining DNAD with self-distillation approaches could eliminate the necessity for a pre-trained model on the target task. Note that it is particularly advantageous when working with novel, task-specific datasets wherever a pre-trained teacher model is not accessible \cite{zhang2021self}. Furthermore, while the current SNPS and DNAD algorithms are primarily designed for image classification tasks, their application to other domains, such as object detection and medical image segmentation \cite{wang2024mednas}, deserve further explorations.

\bibliographystyle{IEEEtran}
\bibliography{bibfile1}

\end{document}